\documentclass[10pt,journal,compsoc]{IEEEtran}
\usepackage{amsmath,amsfonts}
\usepackage{algorithmic}
\usepackage{algorithm}
\usepackage{array}
\usepackage[caption=false,font=normalsize,labelfont=sf,textfont=sf]{subfig}
\usepackage{textcomp}
\usepackage{stfloats}
\usepackage{url}
\usepackage{verbatim}
\usepackage{graphicx}
\usepackage{cite}
\graphicspath{{./pics/}}
\usepackage{multirow}
\usepackage{enumitem}
\usepackage{color}
\usepackage{hyperref}
\newtheorem{definition}{\bf Definition}

\begin{document}

\title{\textit{On Your Mark, Get Set, Predict!} Modeling Continuous-Time Dynamics of Cascades for Information Popularity Prediction}

\author{Xin Jing, Yichen Jing, Yuhuan Lu, Bangchao Deng, Sikun Yang, and Dingqi Yang$^{\ast}$
\thanks{Xin Jing, Yichen Jing, Yuhuan Lu, Bangchao Deng, and Dingqi Yang are with the State Key Laboratory of Internet of Things for Smart City and Department of Computer and Information Science, University of Macau, Macao SAR, China, E-mail: yc27431@um.edu.mo, mc35259@um.edu.mo, yc17462@um.edu.mo, yc37980@um.edu.mo, and dingqiyang@um.edu.mo. Sikun Yang is with the school of computing and information technology at Great Bay University, China, E-mail: sikunyang@gbu.edu.cn}
\thanks{*Corresponding author: Dingqi Yang (email:dingqiyang@um.edu.mo) }
\thanks{Manuscript received April 19, 2021; revised August 16, 2021.}
}

\markboth{Journal of \LaTeX\ Class Files,~Vol.~14, No.~8, August~2021}%
{Shell \MakeLowercase{\textit{et al.}}: A Sample Article Using IEEEtran.cls for IEEE Journals}

\IEEEpubid{0000--0000/00\$00.00~\copyright~2021 IEEE}

\maketitle

\begin{abstract}
Information popularity prediction is important yet challenging in various domains, including viral marketing and news recommendations. The key to accurately predicting information popularity lies in subtly modeling the underlying temporal information diffusion process behind observed events of an information cascade, such as the retweets of a tweet. To this end, most existing methods either adopt recurrent networks to capture the temporal dynamics from the first to the last observed event or develop a statistical model based on self-exciting point processes to make predictions. However, information diffusion is intrinsically a complex continuous-time process with irregularly observed discrete events, which is oversimplified using recurrent networks as they fail to capture the irregular time intervals between events, or using self-exciting point processes as they lack flexibility to capture the complex diffusion process. Against this background, we propose ConCat, modeling the \underline{Con}tinuous-time dynamics of \underline{Ca}scades for information popularity predic\underline{t}ion. On the one hand, it leverages neural Ordinary Differential Equations (ODEs) to model irregular events of a cascade in continuous time based on the cascade graph and sequential event information. On the other hand, it considers cascade events as neural temporal point processes (TPPs) parameterized by a conditional intensity function which can also benefit the popularity prediction task. We conduct extensive experiments to evaluate ConCat on three real-world datasets. Results show that ConCat achieves superior performance compared to state-of-the-art baselines, yielding 2.3\%-33.2\% improvement over the best-performing baselines across the three datasets.
\end{abstract}


\section{Introduction}
\label{Intro}


With the proliferation of information on the Web, the prediction of information popularity in the future is of critical importance~\cite{notarmuzi2022universality}, serving as the foundation of building various applications such as recommendation~\cite{wang2021dydiff}, influence maximization~\cite{li2014efficient}, risk management~\cite{shen2014modeling}, and rumor detection~\cite{lazer2018science}. As typical examples, social networking platforms, such as Facebook\footnote{https://www.facebook.com/} and Twitter\footnote{https://twitter.com/}, incorporate resharing or reposting features that enable users to disseminate content created by others in their online social networks via their friends or followers, in a cascade manner. Such an information dissemination process intrinsically forms an information cascade through the social network~\cite{cheng2014can}. In this context, accurately predicting the popularity of such cascades, which refers to the number of potentially reached users in a future period of time, has attracted increasing attention from both academia and industries, ranging from marketing and traffic control to risk management~\cite{shen2014modeling}.

\begin{figure}[t]\small
	\centering
	\includegraphics[width=0.48\textwidth]{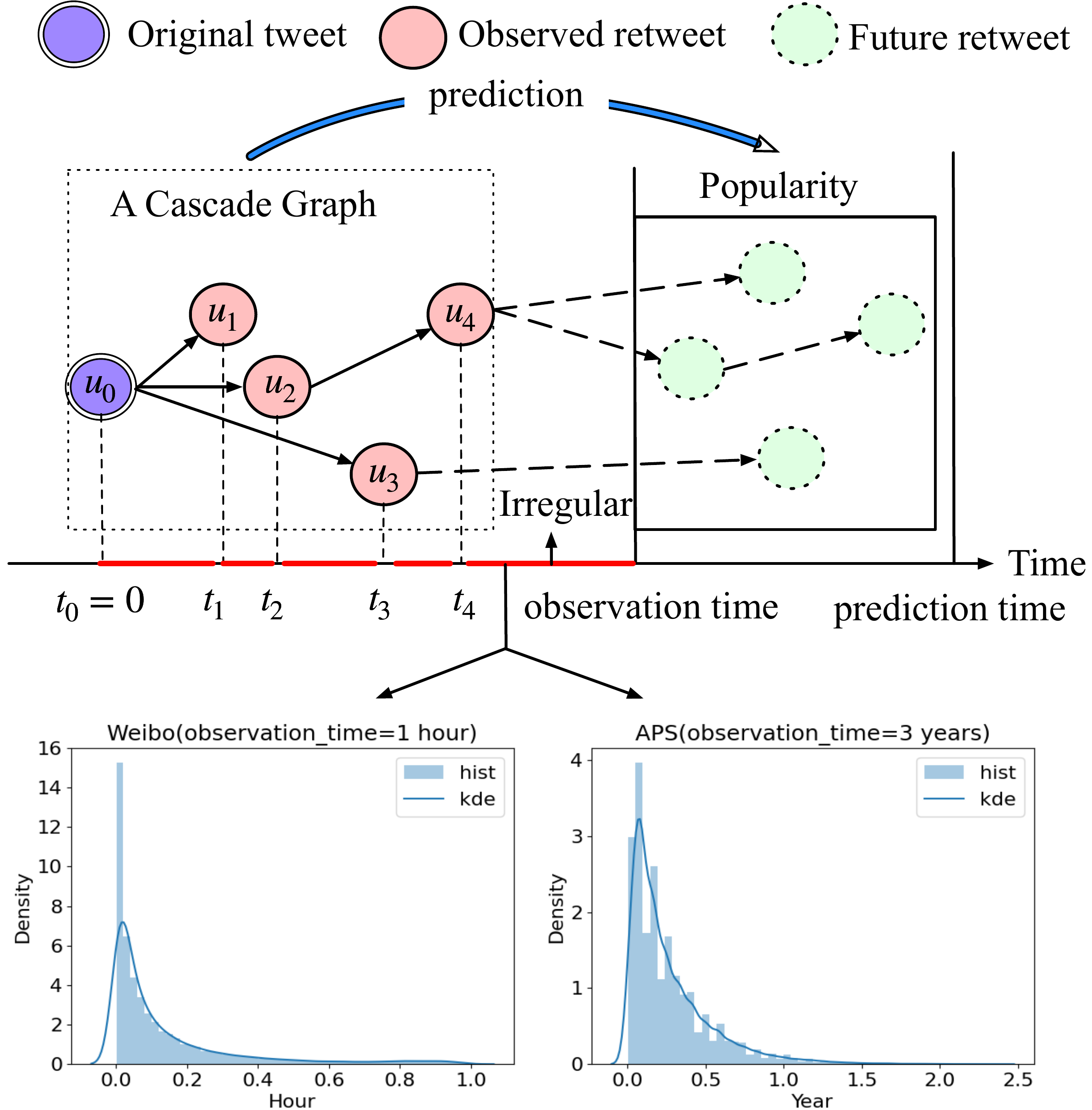}
	\caption{A toy example of a retweet cascade graph (upside) \& Distribution of the irregular time interval between the last node and the observation time of 1 hour and 3 years in Weibo dataset and APS dataset correspondingly (below)}
	\label{Cascade_graph}
\end{figure}

In recent years, many works have focused on understanding the dynamic diffusion process and popularity prediction of cascades~\cite{zhou2021survey, cheng2014can,cao2017deephawkes,xu2021casflow}. These works mainly fall into three categories. First, feature-based approaches manually extract a set of potentially relevant features~\cite{hong2011predicting}, such as publication time~\cite{petrovic2011rt,wu2016unfolding}, user interest~\cite{yang2010understanding}, and then employ various machine learning techniques for prediction, such as regression models~\cite{agarwal2009spatio} and content-based models\cite{naveed2011bad}.
However, these works heavily rely on the quality of hand-crafted features and thus are dataset-specific and have limited generalizability. Second, generative-based approaches resort to generative probabilistic models~\cite{shen2014modeling,zhao2015seismic} for characterizing the cascade diffusion process, but they are built under strong prior assumptions, such as Poisson or Hawkes processes, which are often inflexible to capture the complex diffusion process~\cite{bosser2023predictive}. Finally, deep-learning-based approaches design different deep neural architectures to capture both structural information and temporal dynamics of cascades for information popularity prediction. On the one hand, the structural information of a cascade refers to the interconnections between nodes within a cascade graph. For example, in a retweet cascade graph, as shown in Figure \ref{Cascade_graph}, each retweet event of a user is represented as a node\footnote{As each \textit{node} in a cascade graph represents an \textit{event}, we do not distinguish these two terms throughout this paper.} of the user with an incoming edge from the retweeted user. Existing approaches usually resort to graph embedding techniques to learn node embeddings~\cite{zhou2020variational,chen2019cascn,wang2022casseqgcn} or sample cascade paths as input sequences~\cite{li2017deepcas,tang2021fully} to encode the structural properties of the cascade graph.
On the other hand, the temporal dynamics of the cascade refers to the sequential/temporal patterns of the retweet events in the cascade. Existing approaches mostly regard the learned node embeddings as node features and exploit Recurrent Neural Networks (RNNs), including Long Short-Term Memory (LSTM)~\cite{chen2019cascn,chen2019information}, Gated Recurrent Unit (GRU)~\cite{li2017deepcas,cao2017deephawkes,zhou2020variational,xu2021casflow}, or attention mechanism~\cite{chen2019npp,yu2022transformer} for capturing sequential patterns of the cascade, where the output hidden states are usually used for popularity prediction~\cite{zhou2020variational,chen2019cascn,xu2021casflow}. A few works also combine further temporal features with node features~\cite{cao2017deephawkes,yu2022transformer}, which are together fed to RNNs~\cite{liao2019popularity,chen2019npp}. These deep-learning-based approaches have shown superior performance over feature-based approaches and generative approaches~\cite{cao2017deephawkes,chen2019cascn,zhou2021survey}.

However, the continuous-time dynamics of information cascades intrinsically imply irregularly observed discrete events such as retweets, which are often oversimplified by existing methods. 
Specifically, in a typical prediction setting as shown in Figure \ref{Cascade_graph}, the prediction task takes the input of a cascade graph before an observation time to predict the popularity between the observation time and a prediction time. Two significant issues reside in the current approaches. \textit{First, the continuous-time dynamic and the irregularity of the time intervals between nodes within a cascade are overlooked by simply feeding the node sequences to RNNs~\cite{zhou2020variational,xu2021casflow}}. 
Specifically, these approaches can only output hidden states at discrete timestamps such as $t_0, t_1,$ and $t_2$, etc, and thus utilize the hidden state of the last node ($u_4$ at time $t_4$ as shown in Figure \ref{Cascade_graph}) to forecast the cascade popularity. Here we argue that the hidden state of the last node usually differs from the actual hidden state at the observation time due to their irregular time interval, and it thus can not fully represent the diffusion process. As shown in Figure \ref{Cascade_graph}, we plot the histograms (hist) and its kernel density estimation (kde) of the time interval between the last node and the observation time on Weibo dataset~\cite{cao2017deephawkes} and on APS dataset~\cite{shen2014modeling}, which are widely used benchmarking datasets for information popularity prediction~\cite{cao2017deephawkes,zhou2020variational,xu2021casflow,lu2023continuous}. We clearly observe the irregular and varying time intervals, which are often ignored by existing works. In this paper, we demonstrate that modeling the continuous-time dynamics (e.g., the dynamics between two successive nodes and also after the last nodes) plays an important role in popularity prediction.
\textit{Second, modeling cascades solely based on graphs is limited in capturing the global trend of cascades while generative-based methods, on the other hand, exhibit powerful capabilities in capturing such trends.} However, most generative-based methods are traditional statistical methods under simple prior assumptions, which are less flexible~\cite{zhao2015seismic}. Therefore, it is essential to develop an efficient approach that can integrate generative-based methods with neural networks to enhance the modeling of cascade dynamics. 


Against this background, we propose ConCat, modeling the \underline{Con}tinuous-time dynamics of \underline{Ca}scades for popularity predic\underline{t}ion. 
Specifically, ConCat leverages neural Ordinary Differential Equations (ODEs) to model irregular cascade events in continuous time based on their corresponding graph structure and sequential event information, where neural ODE models continuous patterns between two events, and GRU incorporates jumps into the neural ODE framework for newly observed discrete events, so as to \textit{propagate the hidden state of the last event of different cascades to the unified observation time (i.e., ``On Your Mark'')}. Modeling the continuous-time dynamics captures rich information about the diffusion process and allows us to derive the hidden states for any given time, in particular, the hidden states at exactly the observation time which could significantly improve the popularity prediction performance, as evidenced by our experiments later. Meanwhile, we further enhance the dynamic representation with neural Temporal Point Processes (TPPs), benefiting from the ODE-based representations to compute the conditional intensity function and specify the global trend of cascades by the integral of TPPs. \textit{This integral is combined together with the derived hidden states at exactly the observation time (i.e., ``Get Set'')} for popularity prediction. In summary, our contributions are as follows:


\begin{itemize}[leftmargin=*]
\item{We revisit the existing cascade popularity prediction methods and identify the importance of modeling continuous-time dynamics of cascades for information popularity prediction, in particular, the dynamics after the last event of a cascade until the observation time.}
\item We propose ConCat, a model specifically designed to capture the continuous-time dynamics of cascades for the purpose of popularity prediction. To effectively accommodate the dynamics of cascades, we subtly integrate neural ODEs with GRU to model both continuous and discrete patterns encoded in cascades. Notably, the sequential information of cascades is characterized as the jump condition in GRU, which dedicatedly captures the spikes in cascades. In addition, neural TPPs are utilized to capture the global trend of cascades, thereby further enhancing the prediction performance.
\item We conduct extensive experiments to evaluate ConCat on several real-world information cascade datasets. Results show that ConCat achieves superior performance compared to a sizeable collection of state-of-the-art baselines on the information popularity prediction task, yielding 2.3\%-33.2\% improvement over the best-performing baselines across different datasets. 

\end{itemize}

\section{Preliminaries}
\label{Preliminaries}
In the section, we firstly define the cascade graph, the global graph, and the cascade popularity prediction problem formally with the corresponding notations and then give a brief introduction to neural ODEs and neural TPPs. 

\subsection{Problem Definition}
We take the retweet cascade as an illustrative example as shown in Figure~\ref{Cascade_graph}. However, it's important to note that our work can be easily applied to various modes of information diffusion, such as the Weibo platform, academic publications, and so on. Table~\ref{table-Notations} presents a comprehensive overview of the notations used in this paper.


As shown in Figure~\ref{Cascade_graph}, a Twitter user, denoted as $u_0$, posts a tweet $I$ at time $t_0=0$ (we set the original tweet time to 0). Subsequently, other users can engage with this tweet through various actions, such as commenting, expressing approval through liking, and retweeting. In this paper, we specifically focus on the event of ``retweeting'' which is widely recognized as a significant means of information dissemination on the Twitter social network. After $u_0$ posts the tweet, four other users retweet this tweet. For example, user $u_1$ reposts the tweet from user $u_0$ at time $t_1$, forming a cascade connection as a triplet $(u_0,u_1,t_1)$ and information $I$ diffuses from user $u_0$ to user $u_1$. Subsequent serialized triplets build up a retweet cascade $C$ and the corresponding cascade graph $\mathbf{G}$. Our goal is to predict the future popularity of the information cascade $C$ between the observation time and prediction time, that is the number of triplets in $C$ within this period of time.

Specifically, given an observation time $t_s$, we define the retweet cascade at time $t_s$ as $C(t_s) = \{(u_{k1},u_{k2},t_k)\}_{k\in M}$, where $M$ indicates there are M triplets involved in the diffusion process and $t_k\leq t_s$. In the following, we first give the formal definition of the cascade graph and global graph, followed by the definition of the cascade popularity prediction problem based on these graphs.
\begin{definition}{\textbf{Cascade Graph} --}
Given a tweet $I$ and its corresponding retweet cascade $C(t_s)$ observed at time $t_s$, its cascade graph $\mathbf{G}(t_s)$ is defined as $\mathbf{G}(t_s) = (\mathcal{V}_c(t_s), \mathcal{E}_c(t_s))$, where $\mathcal{V}_c(t_s)$ is the user set of triplets in $C(t_s)$ and $\mathcal{E}_c(t_s)$ is the edge set in the cascade graph where an edge presents that there exists a retweeting action between two nodes.
\end{definition}
\begin{definition}{\textbf{Global Graph} --}
Given all the retweet cascades under the observation time, we define the global graph as $\mathcal{G} = (\mathcal{V}_g, \mathcal{E}_g)$, where the edge in $\mathcal{E}_g$ represents the node relationship from cascading, such as the follower/followee relationship in the social network.
\end{definition}
\begin{definition}{\textbf{Cascade Popularity Prediciton} --}
Given the observed retweet cascade $C(t_s)$ at $t_s$, we predict the incremental popularity ${P} = |C(t_p)| - |C(t_s)|$, where $t_s$ is the observation time, $t_p$ is the prediction time and $|C|$ denotes the number of triplets contained in the cascade $C$.
\end{definition}

\begin{table}[tbp]
	\caption{Notations}
	\label{table-Notations}
	\begin{tabular}{c|c}
		\hline
		{Symbol} &  {Description} \\
		\hline
            $I$ & Information item, such as a tweet or a paper.  \\
            $u$ & User who publishes or retweets a tweet.  \\
		$\mathcal{C}(t)$ & An information cascade at time t.  \\
            $t_s, t_p$ & Observation time, prediciton time. \\
            ${P}$ & Incremental popularity. \\
            $\mathbf{G} = (\mathcal{V}_c,\mathcal{E}_c)$ & A cascade graph with sets of nodes and edges.  \\
            $\mathcal{G} = (\mathcal{V}_g, \mathcal{E}_g)$ & The global graph with sets of nodes and edges.\\
            $\mathbf{A}$, $\mathbf{D}$ & Adjacency matrix and diagonal degree matrix.\\
            ${E_c}(u_i)$ & Embedding of node $u_i$ in cascade graph.\\
            ${E_g}(u_i)$ & Embedding of node $u_i$ in global graph.\\
            $s_i$ & the sequential representation with attention \\
            $\mathbf{h}(t_i)$ & hidden state at any subsequent time in ODE \\
            $\Lambda_{t_{i}}$ & the integral at time $t_i$\\
            $N$ & Number of triplets in one cascade. \\
            $M$ & Number of cascades in the dataset. \\
		\hline
	\end{tabular}
\end{table}

\subsection{Neural Ordinary Differential Equations}
A neural Ordinary Differential Equation (ODE) describes the continuous-time evolution of variables. It represents a transformation of variables over time, where the initial state at time $t_0$, denoted as $\mathbf{h}(t_0)$, is integrated forward using an ODE to determine the transformed state at any subsequent time $t_i$.

\begin{equation}
\label{ODE-1}
\frac{d\mathbf{h}(t)}{dt} = f (\mathbf{h}(t), t; \theta)\quad \text{where} \quad \mathbf{h}(0) = \mathbf{h}_0
\end{equation}
\begin{equation}
\label{ODE-2}
\mathbf{h}(t_i) = \mathbf{h}(t_0) + \int_{t_0}^{t_i}\frac{d\mathbf{h}(t)}{dt} dt,
\end{equation}
where $f$ denotes a neural network, such as a feed-forward or convolutional network, parameterized by $\theta$ to characterize the ODE dynamics.

\subsection{Neural Temporal Point Processes}
A Temporal Point Process (TPP) is a stochastic process that characterizes event sequences of varying lengths within a specified time interval $[0, T]$. A TPP realization can be represented as a sequence of arrival times, denoted as $t = (t_1,...,t_N)$, where $N$ represents the random variable indicating the number of events, and the arrival times are strictly increasing.

The characterization of a TPP involves the definition of the conditional intensity function ${\lambda}^*(t)$ and the calculation of the corresponding log-likelihood.
\begin{equation}
\label{TPP-LL}
log p(\mathbf{t}) = \sum_{i=1}^{N} log{{\lambda}^*(t_i)} - \int_0^T{\lambda}^*(\tau)d\tau
\end{equation}
By minimizing this Negative Log-Likelihood (NLL), the parameters are estimated/learnt to represent the conditional intensity ${\lambda}^*(t)$ to describe the dynamics at different times. In particular, to capture long-range dependency structure, some recent works \cite{shchur2021neural} utilize neural networks to parameterize the conditional intensity of TPPs, and thus these methods are denoted as neural TPPs.
	
\section{ConCat}
\label{Method}
\begin{figure*}[htb]
	\centering
	\includegraphics[width=\textwidth]{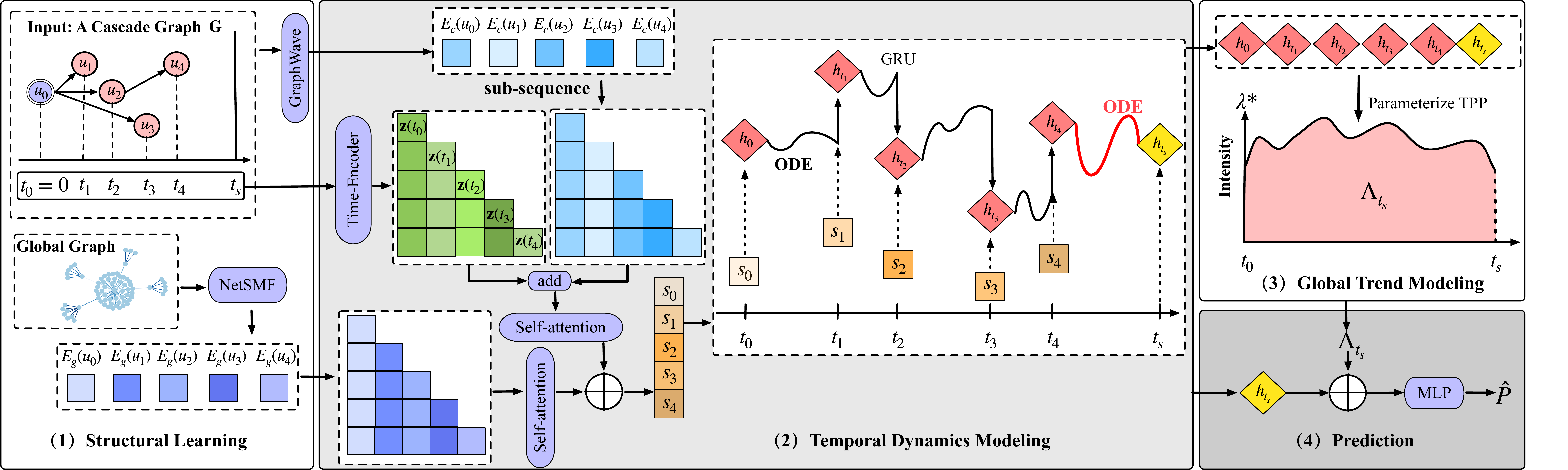}
	\caption{An overview of our proposed model ConCat: (1) The input is a cascade graph $\mathbf{G}$ and the global graph $\mathcal{G}$ for a given observation time and we user GraphWave and NetSMF to model them repsectively, getting the representation $E_c(u_i)$ and $E_g(u_i)$ of each node. (2) We use two self-attention modules to get the sequential information of the cascade graph and global graph, concatenate them to a new vector $\left\{s_{0},s_{1},...\right\}$ and leverage neural ODEs to model the continuous-time dynamics, taking $\left\{s_{0},s_{1},...\right\}$ as the jump condition in GRU. (3) We use the hidden state $h_{t_i}$ to parameterize the TPP and compute the integral $\Lambda_{t_{s}}$. (4) We feed the final hidden state $h_{t_s}$ and $\Lambda_{t_{s}}$ into a MLP for prediciton.   }
	\label{ConCat}
\end{figure*}

In this section, we present the details of our proposed method \textbf{ConCat} to model the \underline{Con}tinuous-time dynamics of \underline{Ca}scades for cascade popularity predic\underline{t}ion, which considers not only the structural information of the cascade graph and global graph but also the irregular continuous-time dynamics characteristic to make information popularity prediction. As shown in Figure~\ref{ConCat}, {ConCat} consists of four main components:

\begin{itemize}[leftmargin=*]
    \item \textbf{(1) Structural learning:} It models and captures contextualized structural characteristics of the cascading graph during diffusion. We use heat wavelet diffusion patterns to learn structural representations of nodes. Meanwhile, we formulate global graph embedding as sparse matrix factorization to efficiently extract node representations from a global view.
    \item \textbf{(2) Temporal dynamics modeling:} Neural ODEs are used to model the continuous information diffusion process. For each newly observed event, we utilize a self-attention mechanism to capture the sequential information from the initial event to the current event and take it as the jump condition in GRU. More importantly, neural ODEs allow us to align the last hidden states of different cascades at a unified observation time (i.e., ``On Your Mark'') for popularity prediction.


    \item \textbf{(3) Global trend modeling:} To capture the global characteristic of the cascades graph, ConCat uses the conditional intensity function to model cascades from the view of TPPs and considers the impact of the integral of intensities for prediction.

    \item \textbf{(4) Prediction:} ConCat combines the aligned hidden states and the integral of the conditional intensity and feeds them into a multi-layer perception (MLP) to make the final popularity prediction.

\end{itemize}



\subsection{Structural Learning}
This component directly uses existing graph embedding methods to model both the cascade graph and the global graph. For the cascade graph, following previous work~\cite{zhou2020variational,xu2021casflow}, we use GraphWave~\cite{donnat2018learning} to capture the local structural information. For the global graph, we employ a fast and scalable network embedding approach NetSMF~\cite{qiu2019netsmf} to obtain connections between all users in the network.

\subsubsection{Cascade Graph Learning}
In our study, we utilize GraphWave~\cite{donnat2018learning} to capture the intrinsic structural characteristics of the cascade graph. This technique enables us to generate a node-level representation by learning the diffusion patterns of spectral graph wavelets for each node. It has been empirically shown as an effective method compared to other graph embedding techniques for cascade popularity prediction problems~\cite{zhou2020variational,xu2021casflow}. Specifically, given the observed cascade graph $\mathbf{G}(t_s)$ at observation time $t_s$, we set the weight between nodes as the time interval of retweet triplet time to observation time and we leverage heat wavelet diffusion patterns to get each node's low-dimensional embeddings $E_c(u_i)$ in the cascade graph, where $u_i \in \mathcal{V}_c$.


\subsubsection{Global Graph Learning}
For the reason that the global graph is a large-scale network, we consider three well-known fast and scalable network embedding methods, including ProNE~\cite{zhang2019prone}, NetSMF~\cite{qiu2019netsmf} and AROPE~\cite{zhang2018arbitrary}. We chose NetSMF to get the representation of the global graph due to its superior performance as evidenced by our experiments later:
\begin{equation}
E_g(u_i) = NetSMF(\mathcal{G}),
\end{equation}
where $\mathcal{G}=(\mathcal{V}_g, \mathcal{E}_g)$ and $u_i \in \mathcal{V}_g$. NetSMF formulates the global graph as a sparse matrix and makes use of spectral sparsification to efficiently sparsify the dense matrix in NetMF~\cite{qiu2018network} with a theoretically bounded approximation error. At last, it performs randomized singular value decomposition to efficiently factorize the sparsified matrix, outputting the node embeddings for the global graph.

\subsection{Temporal Dynamics Modeling}
In this component, to model the irregular temporal dynamics in the cascade graph, we leverage neural ODEs to capture the temporal patterns with real-valued timestamps. However, ODEs cannot incorporate discrete events that abruptly change the latent vector. So we consider a sudden jump conditioned on sub-sequence representation to enhance the continuous dynamic flow.

\subsubsection{Modeling Sequential Information with Self-Attention}


After getting the embeddings $E_c(u)$ of each node in the cascade graph with real-valued timestamps and the embeddings $E_g(u)$ of each node in the global graph without real-valued timestamps, we model the sequential information of the two graphs with two self-attention modules.

Given a cascade graph $\mathbf{G}(t_s)$, we have $|\mathcal{V}_c|$ sequential node embeddings $\mathbf{E_c} = \{E_c(u_0),...,E_c(u_i),...,E_c(u_N)\}_{u_i\in\mathcal{V}_c}$ with the corresponding timestamps $(t_0,t_1,...,t_N)$ and the global graph node embeddings $\mathbf{E_g} = \{E_g(u_0),...,E_g(u_i),...,E_g(u_N)\}_{u_i\in\mathcal{V}_g}$, where $u_N$ is the last node. 

We split one cascade sequence into multiple sub-sequences $\mathbf{S} = \{(u_0),(u_0,u_1),...,(u_0,u_1,...,u_N)\}$ according to chronological order. We use the self-attention~\cite{vaswani2017attention} for each sub-sequence to get the long-term history information. We define a temporal encoding procedure for each timestamp by trigonometric functions:
\begin{equation}
\label{t-encoding}
[\mathbf{z}(t_i)]_j = \left\{
\begin{aligned}
\cos(t_i/10000^{\frac{j-1}{M}}), \text{if j is odd,} \\
\sin(t_i/10000^{\frac{j}{M}}), \text{ if j is even.}
\end{aligned}
\right.
\end{equation}
We deterministically compute $\mathbf{z}(t_i)\in\mathbb{R}^M$, where $M$ is the dimension of encoding. Then, the embedding of each event in the sequence is specified by 
\begin{equation}
\label{event-add}
\mathbf{x}(i) = \mathbf{z}(t_i) + E_c(u_i),
\end{equation}
\begin{equation}
\label{embedding-sequence}
E_{ci}=\{\mathbf{x}(0),\mathbf{x}(1),...,\mathbf{x}(i)\},
\end{equation}
\begin{equation}
\label{embedding-sequence2}
E_{gi}=\{\ E_g(u_0), E_g(u_1),...,E_g(u_i)\},
\end{equation}
where $E_{ci}$ and $E_{gi}$ are the i-th sub-sequence embedding of the cascade graph and global graph respectively and we pass them through two self-attention modules. Specifically, the scaled dot-product attention\cite{vaswani2017attention} is defined as:
\begin{equation}
\label{Softmax}
S = \text{Attention}(Q,K,V) = \text{Softmax}(\frac{QK^T}{\sqrt{d}})V,
\end{equation}
where $Q, K, V$ represent queries, keys, and values. In our case, the self-attention operation takes the embedding $E_{ci}$ and $E_{gi}$ as input, and then converts it into three matrices by linear projections:
\begin{equation}
\label{linear}
Q = EW^Q, K=EW^K, V=EW^V,
\end{equation}
where $W^Q$, $W^K$ and $W^V$ are weights of linear projections. We use a position-wise feed-forward network to transform the attention output $S$ into the hidden representation $s$. For all sub-sequences, we all employ the above self-attentive operation to generate hidden representation ${s_{c0},s_{c1},...,s_{cN}}$ and ${s_{g0},s_{g1},...,s_{gN}}$ for cascade graph and global graph respectively. Then we concatenate the two hidden representations, denoted as ${s_{0},s_{1},...,s_{N}}$, and take it as the jump condition in the dynamic flow.

\subsubsection{Modeling Continuous Dynamics with Neural ODEs}
After getting the jump condition ${s_{0},s_{1},...,s_{N}}$, we leverage neural ODEs to model the dynamics with a vector representation $h_{t_i}$ at every timestamp $t_i$ that acts as both a summary of the past history and as a predictor of future dynamics. Meanwhile, by making instantaneous updates ${s_0,s_1,...,s_N}$ to the hidden state $\mathbf{h}_t$, we can incorporate abrupt changes according to new observed events.

Here we use a standard multi-layer fully connected neural network $f_1$ to model the continuous change in the form of an ODE. When a new retweeting action occurs at time $t_i$, we use a GRU function $g$ to model instantaneous changes based on a newly observed node:

\begin{equation}
\label{ODE-t}
\frac{d{h}_{t_i}}{dt} = f_1 (t, {h}_{t_{i-1}} ) 
\end{equation}
\begin{equation}
\label{ode-solver}
{h}_{t_i}^{'} = \text{ODESolve}(f_1,h_{t_{i-1}},(t_{i-1},t_i))
\end{equation}
\begin{equation}
{h}_{t_i} = g({h}_{t_i}^{'},s_i)
\end{equation}
By solving {equation}~\ref{ode-solver}, we can get a sequence of hidden states $(h_{t_0},h_{t_1},...,h_{t_N})$ after each jump. To propagate the last event of different cascades to the fixed observation time, we also use the ODE to get the hidden representation $h_{t_s}$ exactly at the observation time based on the final output state:
\begin{equation}
{h}_{t_s} = \text{ODESolve}(f_1,h_{t_{N}},(t_{N},t_s))
\end{equation}
We use the hidden representation $h_{t_s}$ as the cascade latent vector for later popularity prediction.

\subsection{Global Trend Modeling}
In this section, we leverage neural temporal point processes to model the probability distribution over variable-length event sequences in continuous time for capturing the global trend in the cascade graph.

Here we use the conditional intensity function $\lambda^*(t)$ to define the temporal point process and parameterize $\lambda^*(t)$ with neural networks. In the cascade graph, we consider the occurring time of every node as the temporal point process $(t_0,t_1,...,t_N)$ and we use the hidden state dynamics ${h}_{t}$ to represent the intensity function:
\begin{equation}
\lambda^*(t) = f_2({h}_{t})
\end{equation}
where $f_2$ is a neural network with a softplus nonlinearity applied to the output, to ensure the intensity is positive. 

Meanwhile, we compute the integral of the conditional intensity $\lambda^*(t)$ based on neural ODE between 0 to $t_s$:
\begin{equation}
\label{lamda}
\frac{d\Lambda_{t_i}}{dt} = f (t, \Lambda_{t_{i-1}} ) \quad \text{where} \quad\Lambda_{t_0} = 0
\end{equation}

We use the ODE dynamics to represent the intensity function because $\Lambda_{t}$ is difficult to compute; most TPPs approximate the integral by using Monte Carlo integration or numerical integration which creates a deviation, while ODE-based methods are closed-formed for computing the integral and neural TPP is an enhancement to the dynamic flow representation.

By solving equation \ref{lamda}, we can get the final integral $\Lambda_{t_s}$. The intensity of TPPs is determined by the distribution of events while high-frequency events represent a large intensity in this period, inspiring us to use the integral of TPPs as a feature to represent the popularity under observation, describing the global trend of the whole diffusion process.

\subsection{Prediction}
After obtaining the final hidden state $h_{t_s}$ at observation time $t_s$ and the integral $\Lambda_{t_s}$ from 0 to $t_s$, we concatenate the integral with the final hidden state and then feed them into an MLP to make the final cascade popularity prediction:
\begin{equation}
\hat{P} = Softplus(MLP([\Lambda_{t_s}, h_{t_s}]))
\end{equation}
where the softplus activation function is to ensure the predicted popularity is positive. The loss function is defined as:

\begin{equation}
\begin{split}
\mathcal{L} =& {({log}_2(P+1) - {log}_2(\hat{P}+1))}^2  \\& - (\sum_{i=1}^{N}log{{\lambda}^*(t_i)} - \int_0^{t_s}{\lambda}^*(\tau)d\tau)
\end{split}
\end{equation}

\begin{equation}
\int_0^{t_s}{\lambda}^*(\tau)d\tau = \Lambda_{t_{s}}
\end{equation}
where $P$ is the truth popularity and $\hat{P}$ is the predicted popularity. The first term is the mean squared logarithmic error, which is a typical loss for regression problems. The second term is the negative log-likelihood of the temporal point process. Note that $\Lambda_{t_{s}}$ is the integral of the conditional intensity function $\lambda^*(t)$ between 0 to $t_s$. 

	
\section{Experiments}
\label{experiments}
In this section, we present an overview of our benchmark datasets and then evaluate our model against state-of-the-art baselines in information cascade popularity prediction task to answer the following questions:
\begin{itemize}[leftmargin=*]
\item {\textbf{RQ1:}} Compared to state-of-the-art baselines, can our approach achieve more accurate prediction for cascade popularity?
\item {\textbf{RQ2:}} What's the impact of different pre-processing settings on performance, such as the number of triplets? 
\item {\textbf{RQ3:}} Why do we need neural ODE to model the irregular cascades? How much does it contribute to performance improvement?
\item {\textbf{RQ4:}} What are the benefits of employing neural TPPs for enhancing the modeling of spatiotemporal dynamics? In contrast, how does the utilization of simpler models such as RNNs capture the global trend?
\item {\textbf{RQ5:}} Which ODESolver should we choose and what is the impact of the hyperparameters?


\end{itemize}

 
 


\subsection{Experimental Settings}
\subsubsection{Datasets \& Pre-processing}
We conduct experiments on several benchmark cascade datasets, considering Sina Weibo~\cite{cao2017deephawkes}, Twitter~\cite{weng2013virality} and APS~\cite{shen2014modeling}, that have been commonly used in previous related works~\cite{chen2019cascn,cao2017deephawkes,shen2014modeling} for evaluating cascade popularity prediction models. Note that the Twitter dataset, originally collected by~\cite{weng2013virality}, consists of publicly available English-written tweets and primarily focuses on examining the impact of network and community structure on diffusion. In this dataset, each hashtag and its corresponding adopters are treated as independent information cascades, but a large number of original tweets' information is missing. For example, user B retweets a tweet from user A at time $t_B$ which is recorded by the dataset, but the time when user A is involved in the hashtag is not available. A significant number of retweeting records, specifically 762,775 out of 1,687,704 (45.2\%), lack the timestamps associated with user A in this example, which prevents us from extracting a real-world cascade information diffusion with sufficient temporal information. \textit{We encourage future works in this direction to refrain from using this Twitter dataset in experiments.} Against this background, we collected a new Twitter dataset crawled from the general Twitter stream. The details of our datasets in experiments are as follows:


\begin{table}
\centering
\caption{Statistics of three datasets}
\begin{tabular}{c c c c}
\hline
Dataset & Twitter & Weibo & APS \\
\hline
Cascades & 86,764 & 119,313 & 207,685 \\
Avg. popularity & 94 & 240 & 51 \\
\hline
\multicolumn{4}{c}{ \emph{Number of cascades in two observation settings}} \\
Train(1d/0.5h/3y) & 7,308 & 21,463 & 18,511 \\
Val(1d/0.5h/3y) & 1,566 & 4,599 & 3,967 \\
Test(1d/0.5h/3y) & 1,566 & 4,599 & 3,966 \\
Train(2d/1h/5y) & 10,983 & 29,908 & 32,102 \\
Val(2d/1h/5y) & 2,353 & 6,409 & 6,879 \\
Test(2d/1h/5y) & 2,353 & 6,408 & 6,879 \\
\hline
\multicolumn{4}{c}{ \emph{Number of nodes and edges in the global graph for the first 100 triplets}} \\
Nodes(1d/0.5h/3y) & 234,522 & 699,279 & 138,695 \\
Edges(1d/0.5h/3y) & 301,353 & 1,040,299 & 275,993 \\
Nodes(2d/1h/5y) & 331,127 & 994,532 & 207,419 \\
Edges(2d/1h/5y) & 446,122 & 1,548,971 & 496,965 \\
\hline
\multicolumn{4}{c}{ \emph{Basic statistics of cascade graphs}} \\
Avg. sequence length & 2.029 & 2.237 & 3.999 \\
Avg. structural virality & 1.872 & 2.025 & 3.114 \\
Avg. page rank & 0.069 & 0.045 & 0.189 \\
Avg. graph density & 0.109 & 0.09 & 0.32 \\
\hline

\end{tabular}
\label{dataset}
\vspace{-0.2cm}
\end{table}

\begin{itemize}[leftmargin=*]
    \item 
    \textit{\textbf{Twitter}} dataset we collected includes the tweets from the general Twitter stream during the period from March 1 to April 15, 2022. We take a hashtag where the original tweet and their retweets with the same hashtag as an information cascade. We select the original tweets during the period from March 1 to March 31, 2022, and their retweets until April 15, ensuring a minimum of 15 days for propagating. 
    \item \textit{Sina \textbf{Weibo}}\footnote{https://github.com/CaoQi92/DeepHawkes} is the largest microblogging platform in China. On Weibo, each original post (also referred to as tweet hereinafter) and its subsequent retweets can generate a retweeting cascade. Due to the diurnal rhythm effect in Weibo~\cite{cao2017deephawkes}, we focus on tweets posted between 8 a.m. and 6 p.m., allowing each tweet at least 6 hours to gather retweets.
    \item \textit{American Physical Society (\textbf{APS})}\footnote{https://journals.aps.org/datasets} contains scientific papers published by APS journals. Every paper in the APS dataset and its citations form a citing cascade. We consider papers published between 1893 and 1997, ensuring a minimum of 20 years (1997 - 2017) for each paper to accumulate citations. 
\end{itemize}

\begin{figure}[t]

        \begin{minipage}{0.48\linewidth}
		\vspace{3pt}
		\centerline{\includegraphics[width=\textwidth]{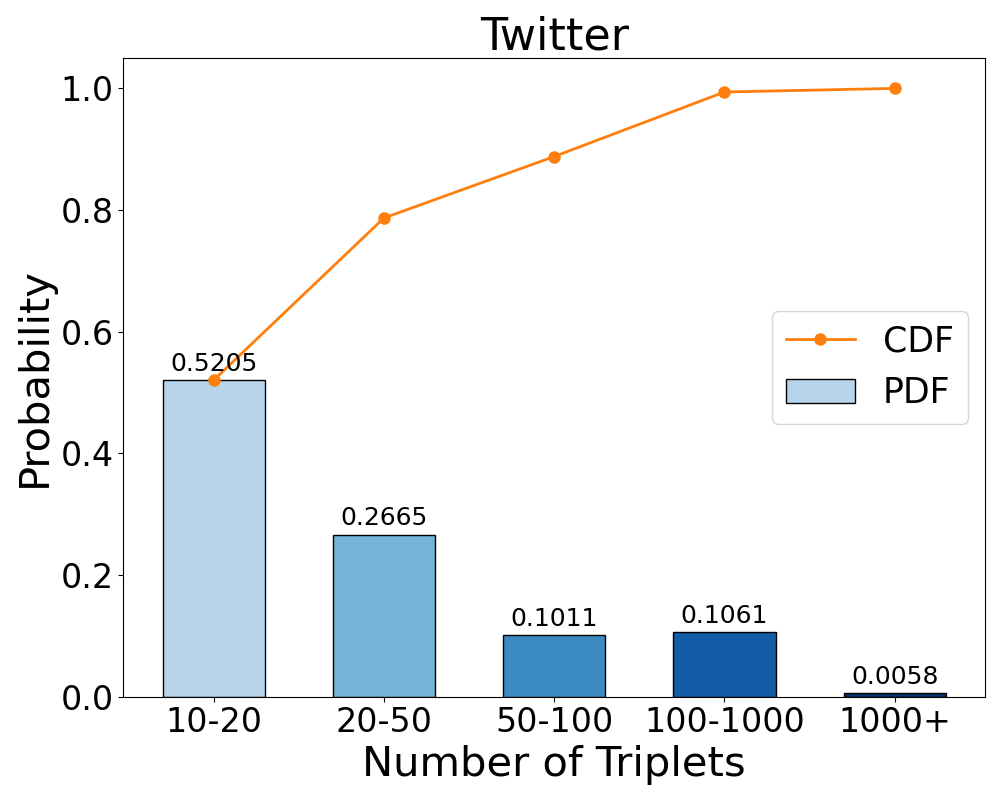}}
	\end{minipage}
	\begin{minipage}{0.48\linewidth}
		\vspace{3pt}
		\centerline{\includegraphics[width=\textwidth]{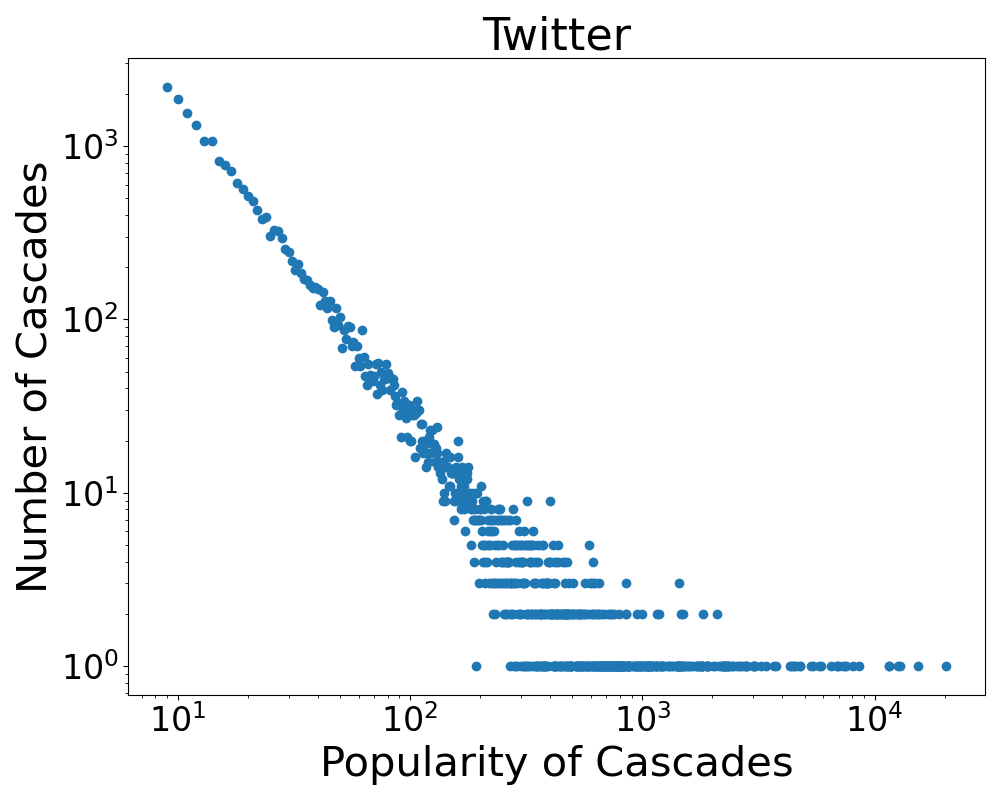}}
	\end{minipage}
 
 	\begin{minipage}{0.48\linewidth}
		\vspace{3pt}
		\centerline{\includegraphics[width=\textwidth]{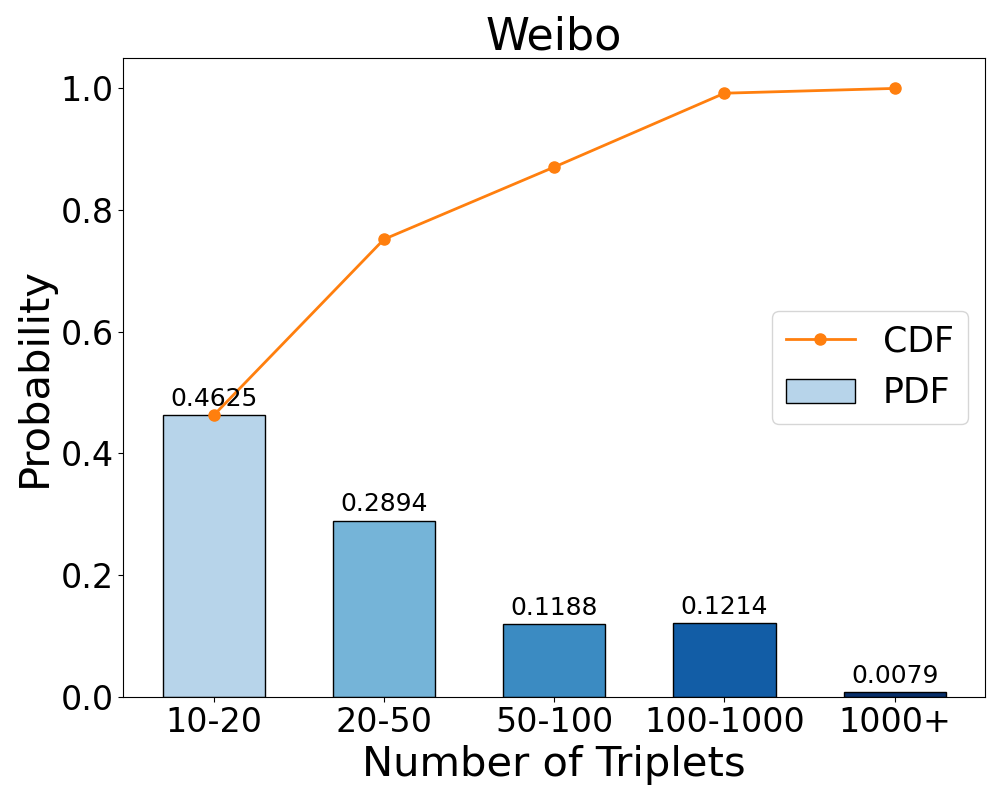}}
	\end{minipage}
	\begin{minipage}{0.48\linewidth}
		\vspace{3pt}
		\centerline{\includegraphics[width=\textwidth]{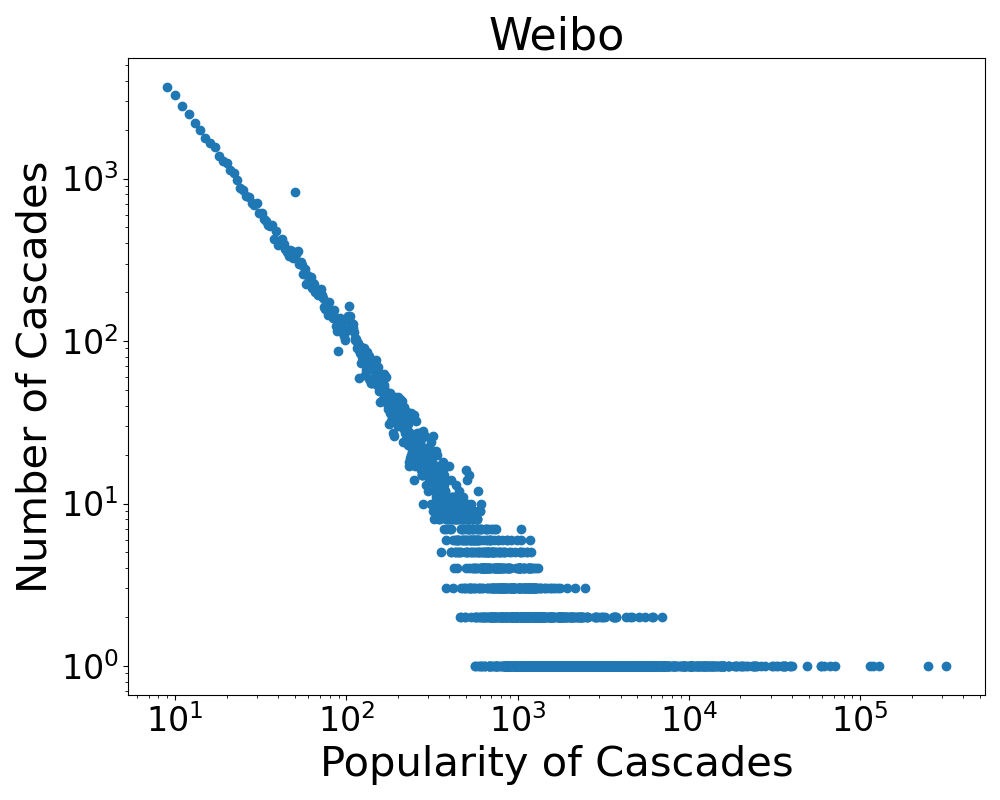}}
	\end{minipage}

        \begin{minipage}{0.48\linewidth}
		\vspace{3pt}
		\centerline{\includegraphics[width=\textwidth]{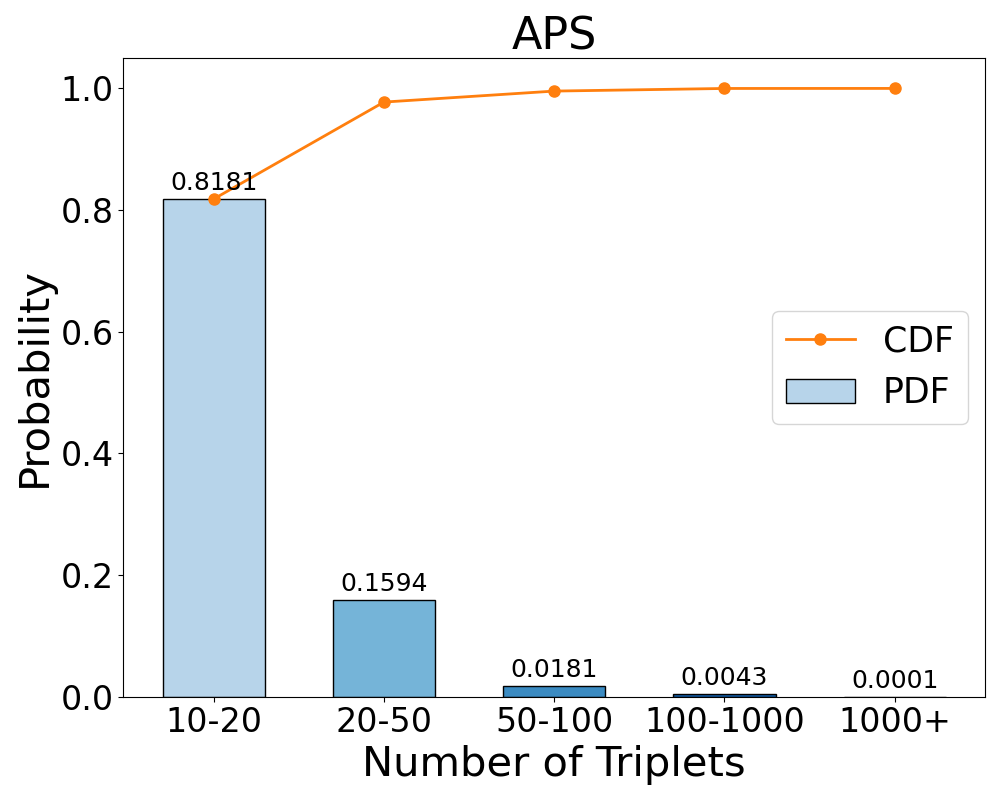}}
	\end{minipage}
	\begin{minipage}{0.48\linewidth}
		\vspace{3pt}
		\centerline{\includegraphics[width=\textwidth]{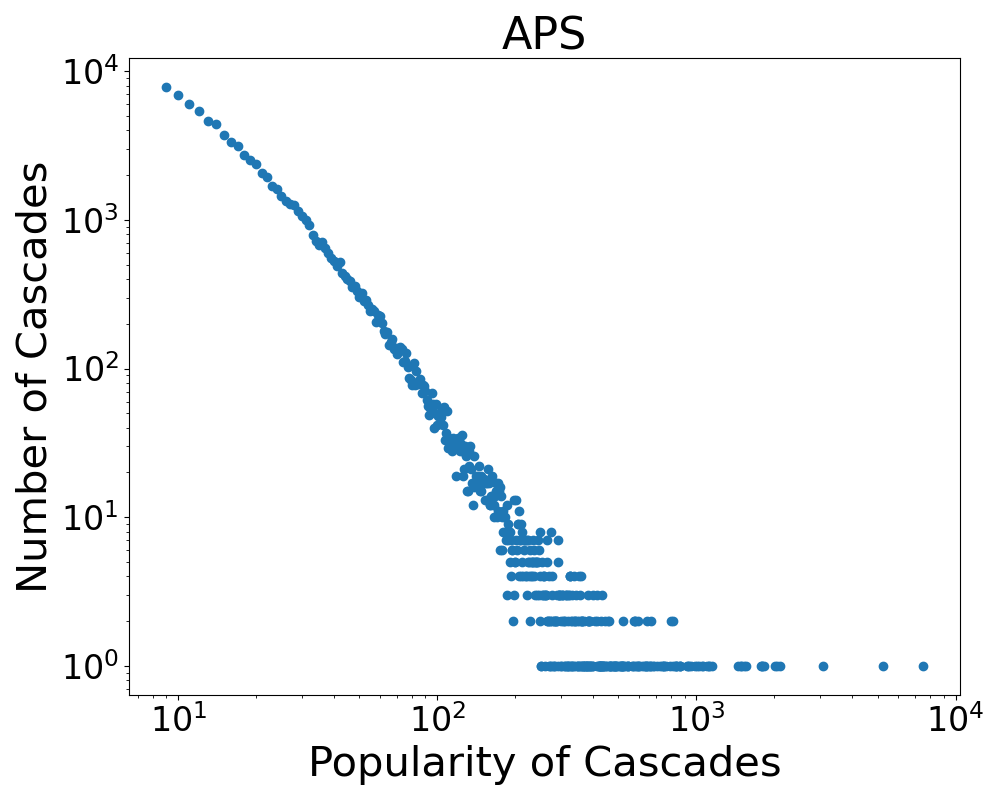}}
	\end{minipage}

	\caption{CDF and PDF of the observed triplets numbers (left) \&  Distribution of popularity (right)}
	\label{Distribution_popularity}
\end{figure}

Following the commonly used settings~\cite{chen2019cascn,xu2021casflow} on these datasets, we set the observation time $t_s$ for Weibo to 0.5 and 1 hour, for APS to 3 and 5 years, and for Twitter to 1 and 2 days. Accordingly, we set the prediction time $t_p$ to 24 hours for Weibo, 20 years for APS, and 15 days for Twitter. Under these settings, we further filter out cascades with fewer than 10 participants at the observation time ($|C(t_s)| < 10$). We consider the first 100 triplets for cascades with more than 100 triplets, consistent with previous work~\cite{zhou2020variational,xu2021casflow,lu2023continuous} for a fair comparison. Moreover, we also expand the number of triplets to further consider temporal dynamics over a longer period of time. We plot the cumulative distribution function (CDF) and the probability density function (PDF) in three datasets under the observation of 0.5 hours, 3 years, and 1 day, respectively, shown in Figure~\ref{Distribution_popularity} (left). We see that there are a large number of cascades where the number of triplets is more than 100, especially in the Weibo and Twitter datasets. We consider the first 1000 triplets consistent as an alternative because we find that triplets for cascades are mostly less than 1000. For all datasets, we split 70\% of the corresponding data for training and the rest for testing (15\%) and validation (15\%). For the global graph, we construct it by the retweeting relationship like~\cite{xu2021casflow} and we only consider the retweeting under the observation time to avoid information leakage. The statistics of the datasets are shown in Table~\ref{dataset}. 

\begin{table*}[tbp]
\centering

\caption{Performance comparison between baselines and ConCat on Twitter datasets under two observation times with different numbers of triplets measured by MSLE, MAPE (lower is better), and $R^2$ ( higher is better)}
\begin{tabular}{l | l  l  l|  l  l  l | l  l l |  l  l  l }
\hline
\multirow{3}{*}{ \textbf{Method} } & \multicolumn{6}{c|}{\textbf{Twitter-100}} & \multicolumn{6}{c}{\textbf{Twitter-1000}} \\
\cline{2-13}
 & \multicolumn{3}{c|}{\textbf{1 day}} & \multicolumn{3}{c|}{\textbf{2 days}}  & \multicolumn{3}{c|}{\textbf{1 day}} & \multicolumn{3}{c}{\textbf{2 days}} \\
\cline{2-13}
 & MSLE & MAPE & $R^2$ & MSLE & MAPE & $R^2$ & MSLE & MAPE & $R^2$ & MSLE & MAPE & $R^2$ \\
\hline
Feature-based &  7.8268 & 0.7073 & 0.1638 &  6.5154 & 0.6514 & 0.1874&  7.3415 & 0.6971 & 0.1764&  6.1474 & 0.6297 & 0.2014\\

SEISMIC &  10.687 & 0.9689 & 0.1148 &  8.1851 & 0.8147 & 0.1431&  10.141 & 0.9581 & 0.1248&  7.4171 & 0.7741 & 0.1644\\

DeepCas &  6.3297 & 0.6566 & 0.2132 &  5.7146 & 0.6671 & 0.2641&  6.1271 & 0.6341 & 0.2347&  5.2471 & 0.6444 & 0.2894\\

DeepHawkes &  5.9341 & 0.5017 & 0.2732  & 4.8489 & 0.5189 & 0.2974&  5.2745 & 0.4587 & 0.3204&  4.4417 & 0.4614 & 0.3477\\

CasCN &  5.8742 & 0.4894 & 0.2831 &  4.7154 & 0.4974 & 0.3041&  
5.2054 & 0.4464 & 0.3341 &  4.4614 & 0.4633 & 0.3549 \\

VaCas &  5.5124 & 0.4796 & 0.2917 &  4.2147 & 0.4871 & 0.3514&  
5.0414 & 0.4396 & 0.3647&  3.9741 & 0.4801 & 0.4315\\

CasFlow  &  \underline{4.7799} & {0.4150} & \underline{0.3739} & {3.6888} & 0.4222 & \underline{0.4846} & \underline{4.1804} & 0.3934 & \underline{0.4437} & \underline{3.0566} &0.3785  & \underline{0.5563}\\

CTCP & {5.3991} & \underline{0.3757} & {0.2815} & \underline{3.6016} &  \underline{0.3773}& {0.4772}& {4.6249} & \underline{0.3730} & {0.3845} & {3.1410} & \underline{0.3608} &{0.5441}\\
\hline
ConCat & \textbf{3.7713} & \textbf{0.3597} & \textbf{0.4981} & \textbf{3.2462} &  \textbf{0.3685}&\textbf{0.5288}& \textbf{3.6587} & \textbf{0.3572} & \textbf{0.5131} & \textbf{2.8007} & \textbf{0.3416} &\textbf{0.5935}\\
\hline
\end{tabular}
\label{Twitter}
\end{table*}

\begin{table*}[tbp]
\centering
\caption{Performance comparison between baselines and ConCat on Weibo datasets under two observation times with different numbers of triplets measured by MSLE, MAPE (lower is better), and $R^2$ ( higher is better)}
\begin{tabular}{l | l  l  l|  l  l  l | l  l l |  l  l  l }
\hline
\multirow{3}{*}{ \textbf{Method} } & \multicolumn{6}{c|}{\textbf{Weibo-100}} & \multicolumn{6}{c}{\textbf{Weibo-1000}} \\
\cline{2-13}
 & \multicolumn{3}{c|}{\textbf{0.5 hour}} & \multicolumn{3}{c|}{\textbf{1 hour}}  & \multicolumn{3}{c|}{\textbf{0.5 hour}} & \multicolumn{3}{c}{\textbf{1 hours}} \\
\cline{2-13}
 & MSLE & MAPE & $R^2$ & MSLE & MAPE & $R^2$ & MSLE & MAPE & $R^2$ & MSLE & MAPE & $R^2$ \\
\hline
Feature-based &  4.0788 & 0.4094 & 0.3914 & 3.6380 & 0.4268 & 0.4389 & 3.3485 & 0.3353 & 0.4157 & 2.9854 & 0.3596 & 0.4294\\

SEISMIC &  5.0300 & 0.4819 & 0.2041 & 4.0594 & 0.5003 & 0.3417& 4.5954 & 0.4824 & 0.4175 & 3.8402& 0.5026 &0.4234\\

DeepCas &  4.6460 & 0.3258 & 0.4314 & 3.5532 & 0.3532 & 0.4514 & 3.7855 & 0.3292 & 0.4716 & 3.2319& 0.3576 &0.4915\\

DeepHawkes &  2.8741 & 0.3041 & 0.5104& 2.7434 & 0.3346 & 0.5264  & 2.3313 & 0.2648 & 0.5977  & 2.2246& 0.2925 & 0.6078 \\

CasCN &  2.7931 & 0.2940 & 0.5341 & 2.6831 & 0.3255 & 0.5415 &  2.2349 & 0.2588 & 0.6141 & 2.1190 & 0.2837 & 0.6207 \\

VaCas &  2.5246 & 0.2847 & 0.5648 & 2.3451 & 0.2997 & 0.5867  & 2.1894 & 0.2544 & 0.6343 & 2.0145 & 0.2748 &  0.6416\\

CasFlow  &  \underline{2.3370} & \underline{0.2665} & \underline{0.5971} & \underline{2.2232} & \underline{0.2949} & \underline{0.6190} & \underline{2.0164} & \underline{0.2509} & \underline{0.6590} & \underline{1.9458} & \underline{0.2632}  &\underline{0.6508}\\

CTCP & 2.5572 & 0.3056 & 0.5733 & 2.2968 & 0.3010 & 0.5934 & 2.1550 & 0.2511 & 0.6404 & 2.0054 & 0.2643 & 0.6450 \\
\hline
ConCat & \textbf{2.1773} & \textbf{0.2457} & \textbf{0.6367} & \textbf{2.0650} & \textbf{0.2695}&\textbf{0.6344} & \textbf{1.8599} & \textbf{0.2315} & \textbf{0.6897} & \textbf{1.8264} & \textbf{0.2521} & \textbf{0.6767}\\
\hline
\end{tabular}
\label{Weibo}
\end{table*}

\begin{table*}[tbp]
\centering
\caption{Performance comparison between baselines and ConCat on APS datasets under two observation times with different numbers of triplets measured by MSLE, MAPE (lower is better), and $R^2$ ( higher is better)}
\begin{tabular}{l | l  l  l|  l  l  l | l  l l |  l  l  l }
\hline
\multirow{3}{*}{ \textbf{Method} } & \multicolumn{6}{c|}{\textbf{APS-100}} & \multicolumn{6}{c}{\textbf{APS-1000}} \\
\cline{2-13}
 & \multicolumn{3}{c|}{\textbf{3 years}} & \multicolumn{3}{c|}{\textbf{5 years}}  & \multicolumn{3}{c|}{\textbf{3 years}} & \multicolumn{3}{c}{\textbf{5 years}} \\
\cline{2-13}
 & MSLE & MAPE & $R^2$ & MSLE & MAPE & $R^2$ & MSLE & MAPE & $R^2$ & MSLE & MAPE & $R^2$ \\
\hline
Feature-based &  1.9881 & 0.3085 & 0.2517 & 1.9696 & 0.3193 & 0.2628 & 2.0059 & 0.2915 & 0.2548 & 1.9762 & 0.3193 &  0.2634\\

SEISMIC &  2.0583 & 0.3013 & 0.2142 & 2.3013 & 0.4320 & 0.2247 & 2.0651& 0.3017 & 0.2215 & 2.3058 & 0.4323 & 0.2278  \\

DeepCas &  2.1051 & 0.2869 & 0.2418 & 1.9260 & 0.3458 &  0.2445 & 2.0780  & 0.2855 & 0.2489 & 1.8815 & 0.3451 & 0.2514 \\

DeepHawkes &  1.9142 & 0.2823 & 0.2671  & 1.8145 & 0.3368 & 0.2817 & 1.8941 & 0.2825 &  0.2704 & 1.7942 & 0.3358 & 0.2845\\

CasCN   & 1.8930 & 0.2763 & 0.2761 & 1.7494 & 0.3208 & 0.2956 &  1.8562 & 0.2784 &  0.2791 & 1.7297 & 0.3241 &0.2984\\

VaCas &  1.7764 & 0.2697 & 0.3014 & 1.6945 & 0.3012 & 0.3417 & 1.7614 & 0.2694 & 0.3111 & 1.6543 & 0.3006 &  0.3498\\

CasFlow  &  \underline{1.4370} & \underline{0.2401} & \underline{0.4573} & \underline{1.3346} & \underline{0.2624} & \underline{0.4781} & \underline{1.4067} & \underline{0.2376} & \underline{0.4612} & \underline{1.2604} &\underline{0.2619}  &\underline{0.5134}\\

CTCP & 1.7676 & 0.3054 & 0.2732 & 1.3751 & 0.2908 & 0.4663 & 1.7066 & 0.3003 &  0.2983 & 1.2749 & 0.2864 & 0.5052 \\
\hline
ConCat & \textbf{1.2529} & \textbf{0.2263} & \textbf{0.4848} & \textbf{1.1733} &  \textbf{0.2566}&\textbf{0.5446}& \textbf{1.2459} & \textbf{0.2283} & \textbf{0.4877} & \textbf{1.1585} & \textbf{0.2520} & \textbf{0.5504}\\
\hline
\end{tabular}
\label{APS}
\end{table*}

\subsubsection{Baselines}
We select eight state-of-the-art baselines for comparison as follows:

\begin{itemize}[leftmargin=*]
    \item \textbf{Feature-based} approaches extract various hand-crafted features and here we use the following features: the size of the observed cascade, the temporal interval between the original node and its initial forwarding, the time at which the last retweet occurs, and the average duration of the diffusion process from the initial node to the ultimate node like~\cite{xu2021casflow}. These features are subsequently fed into an MLP model for the purpose of predicting the popularity of cascades.
    \item \textbf{SEISMIC}~\cite{zhao2015seismic} develops a no-training and no-expensive feature engineering framework which is based on the theory of self-exciting point processes.
    \item \textbf{DeepCas}~\cite{li2017deepcas} introduces a novel approach to representing cascades which are conceptualized as a collection of random walk paths and it employs a bi-directional GRU with an attention mechanism to effectively model and predict cascades.
    \item \textbf{DeepHawkes}~\cite{cao2017deephawkes} integrates the Hawkes process and deep learning techniques for the purpose of modeling cascades. It takes into account three important factors from the perspective of the Hawkes process: the impact of users, the self-exciting mechanism, and the time decay effects.
    \item \textbf{CasCN}~\cite{chen2019cascn} samples a cascade graph by organizing it into a sequence of sub-cascade graphs. It effectively captures the directions of diffusion and the timing of retweeting through the utilization of graph convolutional networks and LSTM.
    \item \textbf{VaCas}~\cite{zhou2020variational} proposes a hierarchical graph learning method and it leverages variational autoencoder for modeling the uncertainty between the sub-graphs and the whole cascades.
    \item \textbf{CasFlow}~\cite{xu2021casflow} is the extension of VaCas. It mainly considers the effect of both the local graph and global graphs and leverages scalable graph learning methods to represent user behavior for predicting popularity.
    \item \textbf{CTCP}~\cite{lu2023continuous} combines all cascades into a diffusion graph and takes the correlation between cascades and the dynamic preferences of users into account. It proposes an evolution learning module that updates the states of users and cascades in real time as diffusion behaviors occur.
\end{itemize} 

\subsubsection{Metrics}
The distribution of popularity in the three datasets is shown in Figure~\ref{Distribution_popularity} (right) and we see they are all a long-tail distribution where it is better to use log-level metrics for evaluation. We use three commonly used metrics, i.e., mean squared logarithmic error (MSLE), mean absolute percentage error (MAPE), and the coefficient of determination ($R^2$) for evaluation, which are defined as follows:
\begin{equation}
MSLE = \frac{1}{M}\sum_{k=1}^{M}({log}_2(P+1) - {log}_2(\hat{P}+1))^2
\end{equation}
\begin{equation}
MAPE = \frac{1}{M}\sum_{k=1}^{M}{\frac{|{log}_2(P+2) - {log}_2(\hat{P}+2)|}{{log}_2(P+2)}} 
\end{equation}
\begin{equation}
R^2 = 1 - \frac{\sum_{k=1}^{M}({log}_2(P+1) - {log}_2(\hat{P}+1))^2}
{\sum_{k=1}^{M}
({log}_2(P+1) - \frac{1}{M}\sum_{k=1}^{M}{log}_2(P+1))^2}
\end{equation}
where the '+ 1' and the '+ 2' operations are for scale, M is the number of cascades, $P$ is the truth popularity, and $\hat{P}$ is the predicted popularity. Note that the scale is important to avoid the value of zero in logarithmic calculations and denominators. We keep all the baselines to the same scale for evaluation. A smaller value of MSLE and MAPE implies a better performance while a higher $R^2$ means better.

\subsection{Performance Comparison (RQ1\&RQ2)}
Tables~\ref{Twitter}, \ref{Weibo}, and \ref{APS} show the results on the three datasets, respectively. We observe that ConCat significantly outperforms all baselines on MSLE, MAPE, and $R^2$. For example, compared to the best-performing baselines, our ConCat achieves a 6.1\%-21.1\% improvement of MSLE, a 2.2\%-7.8\% improvement of MAPE, and a 2.5\%-33.2\% improvement of $R^2$ on three datasets with two different observation times.
Moreover, we see that the performance of considering the first 1000 triplets is better than the case of 100, especially in Weibo and Twitter datasets, as longer cascades usually provide more information for popularity prediction while there is only a slight difference in the APS dataset because the proportion of cascades with more than 100 triplets is very small (only 0.01\%). Moreover, with the increase in observation time, the MSLE exhibits a decreasing trend while the MAPE shows a larger magnitude. This can be attributed to the fact that as the observation time increases, the incremental popularity diminishes, leading to a smaller denominator in the MAPE calculation and consequently resulting in a larger MAPE value. 



\begin{figure}[t]\small
    \centering
    \includegraphics[width=0.46\textwidth]{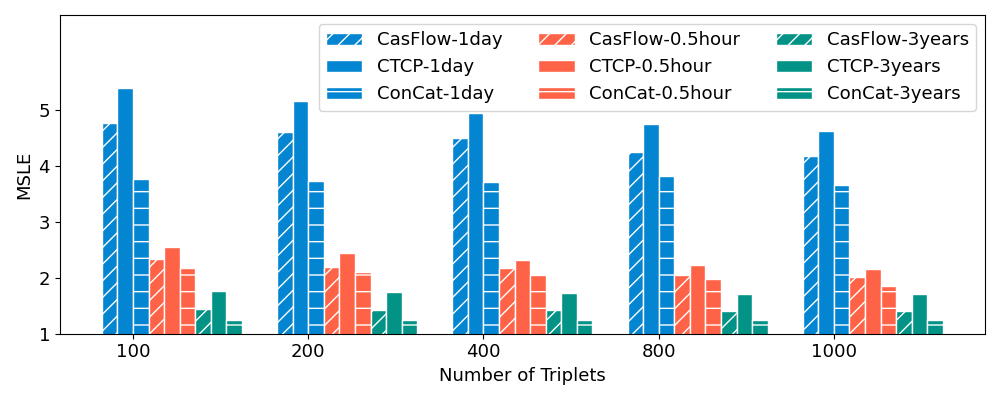}
    \caption{Impact of the number of triplets. We set the number of triplets to 100, 200, 400, 800, and 1000 on three datasets, and we compare ConCat with the top two baselines CasFlow and CTCP.}
    \label{triplets}
\end{figure}

To explore the impact of triplets on the results of the prediction of popularity, we also set the number of triplets to 100, 200, 400, 800, 1000. The results are shown in Figure~\ref{triplets}. The MSLE of Weibo and Twitter demonstrates a decreasing trend as the number of observed triplets increases. The MSLE of APS tends to be stable, as the proportion of cascades with a large number of triplets in Weibo and Twitter significantly surpasses that in APS. 



\begin{table*}[bp]
\centering
\caption{Performance comparison between ConCat and ConCat-variants on Twitter datasets under two observation times with different numbers of triplets measured by MSLE, MAPE (lower is better), and $R^2$ ( higher is better)}
\begin{tabular}{l | l  l  l|  l  l  l | l  l l |  l  l  l }
\hline
\multirow{3}{*}{ \textbf{Method} } & \multicolumn{6}{c|}{\textbf{Twitter-100}} & \multicolumn{6}{c}{\textbf{Twitter-1000}} \\
\cline{2-13}
 & \multicolumn{3}{c|}{\textbf{1 day}} & \multicolumn{3}{c|}{\textbf{2 days}}  & \multicolumn{3}{c|}{\textbf{1 day}} & \multicolumn{3}{c}{\textbf{2 days}} \\
\cline{2-13}
 & MSLE & MAPE & $R^2$ & MSLE & MAPE & $R^2$ & MSLE & MAPE & $R^2$ & MSLE & MAPE & $R^2$ \\
\hline
ConCat-w/o TPP & 4.0324 & 0.3741 & 0.4667 & {3.4588} & {0.3849} & 0.4915 & {3.8745} & {0.3702} & {0.4933} & {3.0514} & {0.3748}  & 0.5550 \\
ConCat-w/o align & 4.1744 & 0.3897 & 0.4557 & {3.5337} & {0.3989} & 0.4877 & {3.9664} & {0.3814} & {0.4871} & {3.1541} & {0.3877}  & 0.5364\\
ConCat-w/o ODE & 4.8641 & 0.3964 & 0.4122 & {4.1154} & {0.4531} & 0.4667 & {4.2341} & {0.3899} & {0.4467} & {3.9947} & {0.4433}  & 0.4560\\
ConCat-w/o attention & 3.8214 & 0.3612 & 0.4862 & {3.2987} & {0.3711} & 0.5202 & {3.7221} & {0.3602} & {0.4996} & {2.8776} & {0.3498}  & 0.5902\\
\hline
ConCat-RNNs & 5.0771 & 0.4157 & 0.3620 & {4.2547} & {0.4364} & 0.3841 & {4.8974} & {0.4054} & {0.3334} & {3.8741} & {0.4215}  & 0.3995\\
ConCat-ELP & 3.8994 & 0.3662 & 0.4799 & {3.3461} & {0.3769} & 0.5099 & {3.7441} & {0.3648} & {0.4821} & {2.9357} & {0.3611}  & 0.5864\\

ConCat-AROPE & 4.2433 & 0.3850 & 0.4220 & {3.4918} & {0.3996} & 0.4786 & {4.0801} & {0.3704} & {0.4430} & {3.0324} & {0.3614}  & 0.5598\\

ConCat-ProNE & 4.0843 & 0.3680 & 0.4432 & {3.3294} & {0.3810} & 0.5167 & {3.7976} & {0.3601} & {0.4946} & {2.8888} & {0.3599}  & 0.5807\\
\hline
ConCat & \textbf{3.7713} & \textbf{0.3597} & \textbf{0.4981} & \textbf{3.2462} &  \textbf{0.3685}&\textbf{0.5288}& \textbf{3.6587} & \textbf{0.3572} & \textbf{0.5131} & \textbf{2.8007} & \textbf{0.3416} &\textbf{0.5935}\\
\hline
\end{tabular}
\label{Twitter_variants}
\end{table*}

\begin{table*}[bp]
\centering
\caption{Performance comparison between ConCat and ConCat-variants on Weibo datasets under two observation times with different numbers of triplets measured by MSLE, MAPE (lower is better), and $R^2$ ( higher is better)}
\begin{tabular}{l | l  l  l|  l  l  l | l  l l |  l  l  l }
\hline
\multirow{3}{*}{ \textbf{Method} } & \multicolumn{6}{c|}{\textbf{Weibo-100}} & \multicolumn{6}{c}{\textbf{Weibo-1000}} \\
\cline{2-13}
 & \multicolumn{3}{c|}{\textbf{0.5 hour}} & \multicolumn{3}{c|}{\textbf{1 hour}}  & \multicolumn{3}{c|}{\textbf{0.5 hour}} & \multicolumn{3}{c}{\textbf{1 hours}} \\
\cline{2-13}
 & MSLE & MAPE & $R^2$ & MSLE & MAPE & $R^2$ & MSLE & MAPE & $R^2$ & MSLE & MAPE & $R^2$ \\
\hline
ConCat-w/o TPP & 2.3418 & 0.2598 & 0.6041 & {2.2644} & {0.2847} & 0.6115 & {1.9840} & {0.2391} & {0.6731} & {1.9830} & {0.2671} & 0.6525\\
ConCat-w/o align & 2.3512 & 0.2604 & 0.6083 & {2.2940} & {0.2899} & 0.6066 & {1.9714} & {0.2384} & {0.6654} & {1.9343} & {0.2611} & 0.6513\\
ConCat-w/o ODE & 2.3964 & 0.2745 & 0.6001 & {2.3412} & {0.2933} & 0.6010 & {2.1945} & {0.2598} & {0.6102} & {2.1735} & {0.2835} & 0.6086\\
ConCat-w/o attention & 2.2347 & 0.2498 & 0.6217 & {2.1340} & {0.2713} & 0.6241 & {1.9947} & {0.2378} & {0.6716} & {1.8946} & {0.2587} & 0.6548\\
\hline
ConCat-RNNs & 2.4487 & 0.2865 & 0.5876 & {2.3531} & {0.2939} & 0.5977 & {2.2014} & {0.2680} & {0.6024} & {2.1840} & {0.2941} & 0.6052\\
ConCat-ELP & 2.3041 & 0.2534 & 0.6188 & {2.1874} & {0.2746} & 0.6204 & {1.9404} & {0.2361} & {0.6547} & {1.9641} & {0.2633} & 0.6411\\
ConCat-AROPE & 2.3669 & 0.2655 & 0.6051 & {2.1293} & {0.2704} & 0.6231 & {1.9560} & {0.2337} & {0.6736} & {1.8493} & {0.2536} & 0.6726\\
ConCat-ProNE & 2.2875 & 0.2522 & 0.6183 & {2.0912} & {0.2695} & 0.6298 & {1.9106} & {0.2317} & {0.6812} & {1.9846} & {0.2917} & 0.6310\\
\hline
ConCat & \textbf{2.1773} & \textbf{0.2457} & \textbf{0.6367} & \textbf{2.0650} & \textbf{0.2695}&\textbf{0.6344} & \textbf{1.8599} & \textbf{0.2315} & \textbf{0.6897} & \textbf{1.8264} & \textbf{0.2521} & \textbf{0.6767}\\
\hline
\end{tabular}
\label{Weibo_variants}
\end{table*}

\begin{table*}[tbp]
\centering
\caption{Performance comparison between ConCat and ConCat-variants on APS datasets under two observation times with different numbers of triplets measured by MSLE, MAPE (lower is better), and $R^2$ ( higher is better)}
\begin{tabular}{l | l  l  l|  l  l  l | l  l l |  l  l  l }
\hline
\multirow{3}{*}{ \textbf{Method} } & \multicolumn{6}{c|}{\textbf{APS-100}} & \multicolumn{6}{c}{\textbf{APS-1000}} \\
\cline{2-13}
 & \multicolumn{3}{c|}{\textbf{3 years}} & \multicolumn{3}{c|}{\textbf{5 years}}  & \multicolumn{3}{c|}{\textbf{3 years}} & \multicolumn{3}{c}{\textbf{5 years}} \\
\cline{2-13}
 & MSLE & MAPE & $R^2$ & MSLE & MAPE & $R^2$ & MSLE & MAPE & $R^2$ & MSLE & MAPE & $R^2$ \\
\hline
ConCat-w/o TPP & 1.3046 & 0.2304 & 0.4607 & {1.1864} & {0.2684} & 0.5301 & {1.3010} & {0.2361} & {0.4812} & {1.1891} & {0.2664} & 0.5312\\
ConCat-w/o align & 1.3144 & 0.2364 & 0.4514 & {1.2041} & {0.2712} & 0.5244 & {1.3151} & {0.2354} & {0.4731} & {1.2020} & {0.2701} & 0.5297\\
ConCat-w/o ODE & 1.4147 & 0.2504 & 0.4308 & {1.3544} & {0.2805} & 0.4494 & {1.4167} & {0.2501} & {0.4319} & {1.3541} & {0.2801} & 0.4403\\
ConCat-w/o attention & 1.2504 & 0.2234 & 0.4807 & {1.2007} & {0.2534} & 0.5240 & {1.2514} & {0.2294} & {0.4897} & {1.1974} & {0.2511} & 0.5307\\
\hline
ConCat-RNNs & 1.4760 & 0.2514 & 0.4245 & {1.4143} & {0.2817} & 0.4379 & {1.4667} & {0.2508} & {0.4262} & {1.4013} & {0.2754} & 0.4398\\
ConCat-ELP & 1.2848 & 0.2297 & 0.4784 & {1.1773} & {0.2655} & 0.5353 & {1.2828} & {0.2330} & {0.4749} & {1.1852} & {0.2650} & 0.5332\\
ConCat-AROPE & 1.3156 & 0.2369 & 0.4591 & {1.2349} & {0.2625} & 0.5207 & {1.3357} & {0.2365} & {0.4508} & {1.2165} & {0.2610} & 0.5279\\

ConCat-ProNE & 1.2614 & 0.2271 & 0.4814 & {1.1973} & {0.2555} & 0.5353 & {1.2528} & {0.2315} & {0.4849} & {1.1625} & {0.2554} & 0.5488\\
\hline
ConCat & \textbf{1.2529} & \textbf{0.2263} & \textbf{0.4848} & \textbf{1.1733} &  \textbf{0.2566}&\textbf{0.5446}& \textbf{1.2459} & \textbf{0.2283} & \textbf{0.4877} & \textbf{1.1585} & \textbf{0.2520} & \textbf{0.5504}\\
\hline
\end{tabular}
\label{APS_variants}
\end{table*}

\subsection{Ablation Study (RQ3\&RQ4)}
To answer RQ3 and RQ4, we have conducted a series of experiments where we have introduced different variants of our ConCat model. The purpose of these variants is to examine the impact of key design choices on the model's performance. Specifically, we investigate: 1) whether it is useful to model the continuous-time dynamics of cascades utilizing neural ODEs instead of RNNs, 2) whether it is useful to propagate the last event of different cascades to the unified observation time (i.e., “On Your Mark”) for popularity prediction, 3) whether it is helpful to capture the global trend of the cascade graph by integrating neural TPPs rather than directly leveraging the last hidden state in temporal modeling for prediction, and 4) whether it is necessary to employ neural TPPs to capture the global trend, as opposed to some simpler methods. 

%

Firstly, we conduct the ablation study to investigate the contribution of each component as follows:
\begin{itemize}[leftmargin=*]
    \item \textbf{ConCat-w/o TPP}: We remove the global trend modeling module and only input the hidden state $h_{t_s}$ to MLP. The loss function only considers MSLE without the log-likelihood.

    \item \textbf{ConCat-w/o align}: We remove the hidden state $h_{t_s}$ and only rely on the last node's hidden representation to make predictions.

    \item \textbf{ConCat-w/o ODE}: We remove the ODE module and it is correspondingly without the global trend modeling because the integral is computed by ODE.

    \item \textbf{ConCat-w/o attention}: We remove the self-attention block, concatenate the representation of the cascade graph and global graph, and take it as the jump condition.

\end{itemize}

Secondly, we design some variants to better answer our design choice in our model:

\begin{itemize}[leftmargin=*]
    \item \textbf{ConCat-RNNs}: We remove the ODE and TPP modules, concatenate the representation of the cascade graph and global graph, and feed it into two bi-directional GRUs according to the time order.

    \item \textbf{ConCat-ELP}: We replace the global trend modules with another module, which separates the observation time into five equal-length periods and counts the popularity of a cascade in each time interval. Then we feed the series of the popularity count into an MLP to get global trend representation and concatenate it with the hidden state $h_{t_s}$ for prediction.

    \item \textbf{ConCat-AROPE \& ConCat-ProNE}: We replace the global graph learning method with other scalable network embedding approaches, such as ProNE~\cite{zhang2019prone} and AROPE~\cite{zhang2018arbitrary}.

\end{itemize}

The results and comparison of these variants are shown in Table~\ref{Twitter_variants}, Table~\ref{Weibo_variants} and~\ref{APS_variants}, on the three datasets, respectively.
First, compared to ConCat-RNNs and ConCat-w/o ODE, ConCat shows a significant improvement of 9.14\%-29.89\% on MSLE across all datasets, showing the effectiveness of modeling the continuous-time dynamics of cascades for popularity prediction. Additionally, We can see the attention module has enhanced the dynamic representation by comparing ConCat-w/o attention with ConCat.
Second, compared to ConCat-w/o align, ConCat shows 2.56\%-11.20\% improvement on MSLE, showing the usefulness of propagating the last event to the observation time. 
Third, the comparison of ConCat over ConCat-w/o TPP also validates the utility of capturing the global trend of cascade using neural TPP with an improvement of up to 8.81\%.
Fourth, through a comparative analysis of ConCat-ELP and ConCat-w/o TPP, it becomes evident that ConCat-ELP exhibits superior performance across various metrics, implying the importance of incorporating global trends in the prediction process. Moreover, ConCat further surpasses ConCat-ELP in all scenarios, showing the superiority of neural TPPs over simple methods in capturing such global trends.
Finally, we compare the performance of different global graph embedding methods and choose NetSMF as our global graph representation approach for better results.

\subsection{Hyperparameter Sensitivity (RQ5)}
\begin{figure}[t]
	\begin{minipage}[t]{0.32\linewidth}
		\centerline{\includegraphics[width=\textwidth]{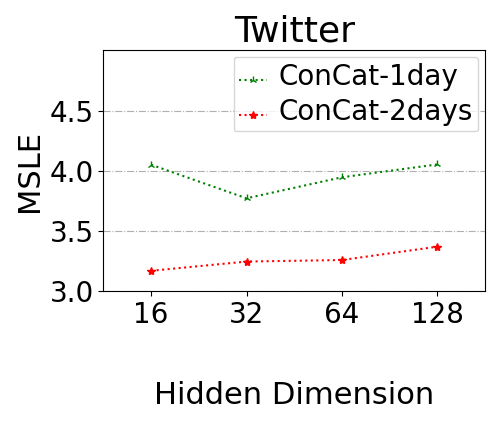}}
	\end{minipage}
	\begin{minipage}[t]{0.32\linewidth}
		\centerline{\includegraphics[width=\textwidth]{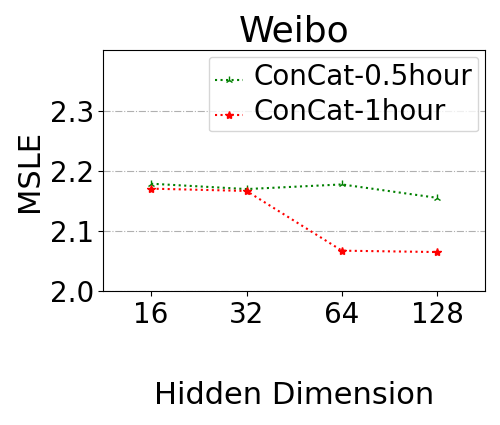}}
	\end{minipage}
	\begin{minipage}[t]{0.32\linewidth}
		\centerline{\includegraphics[width=\textwidth]{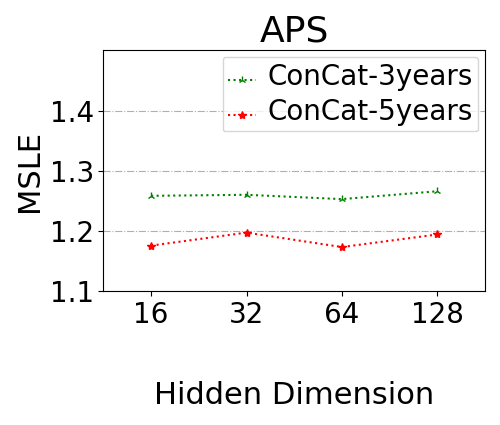}}
	\end{minipage}
	\caption{Impact of the hidden dimension of ${h}_{t}$ for first 100 triplets.}
        \label{Hyperparameters}
\end{figure}

We investigate the influence of hyperparameters including the hidden dimension of ${h}_{t}$ and the choice of ODESolvers. The hidden dimension is varied across values of 16, 32, 64, and 128, while the ODESolver options considered are dopri5, bosh3, adaptive\_heun, euler, rk4, implicit\_adams, and midpoint. The results of the impact of different hidden dimensions are shown in Figure~\ref{Hyperparameters} and we see that the hidden dimensions have some influence on the Weibo and Twitter datasets while the performance on APS indicates a tendency toward stability. 

Additionally, among the ODESolver options shown in Table~\ref{hyper_para_odesolver}, euler yields the worst performance because the error of euler's method usually decreases as the step size decreases, while dopri5 demonstrates relatively superior predictive capabilities. Therefore, we empirically select the best hyperparameters for individual datasets in previous experiments, choose the dopri5 as our ODESolver across all of our experiments, and set the hidden dimension of ${h}_{t}$ as 32 for Twitter, 64 for APS and Weibo.

\begin{table}[tbp]\scriptsize
\centering
\caption{Impact of different types of ODE\_Solver on Twitter (1 day), Weibo (0.5 hour), and APS (3 years) for first 100 triplets.}
\begin{tabular}{l | l  l | l l | l l }
\hline
\multirow{2}{*}{ \textbf{ODE\_Solver} } & \multicolumn{2}{c|}{Twitter} & \multicolumn{2}{c|}{Weibo} & \multicolumn{2}{c}{APS} \\
\cline{2-7}
 & MSLE & MAPE & MSLE & MAPE & MSLE & MAPE\\
\hline

bosh3 & 4.0821 & 0.3753 & 2.1792 & 0.2509 & 1.2701 & 0.2292 \\
adaptive\_heun & 4.2103 & 0.3722 & 2.1786 & 0.2578 & 1.2702 & 0.2291 \\
euler & 4.2396 & 0.3983 & 2.2188 & 0.2598 & 1.2837 & 0.2304 \\
rk4 & 4.2005 & 0.3978 & 2.1780 & 0.2537 & 1.2644 & 0.2263 \\
implicit\_adams & 4.0612 & 0.3710 & 2.1780 & 0.2537 & 1.2644 & 0.2263 \\
midpoint & 3.9759 & 0.3679 & 2.1873 & 0.2457 & 1.2622 & 0.2280 \\
dopri5 & \textbf{3.9459} & \textbf{0.3679} & \textbf{2.1773} & \textbf{0.2457} & \textbf{1.2529} & \textbf{0.2263} \\
\hline
\end{tabular}
\label{hyper_para_odesolver}
\end{table}

\section{Related Work}
\label{RelatedWork}


In the current literature, cascade popularity prediction techniques can be roughly classified into three categories.

\textbf{Feature-based approaches}: In the early stage of this research direction, the primary emphasis was placed on understanding and identifying diverse hand-crafted features derived from raw data. These features include cascade graph structures ~\cite{gao2014effective}, temporal features such as publication time~\cite{petrovic2011rt,wu2016unfolding}, observation time~\cite{cheng2014can,yang2011patterns}, first participation time~\cite{zaman2014bayesian}, and user interest~\cite{yang2010understanding}, contents~\cite{hong2011predicting}, etc. After the extraction of these features, they are then fed into conventional machine learning models for popularity prediction, such as simple regression models~\cite{agarwal2009spatio}, regression trees~\cite{bakshy2011everyone}, support vector machine~\cite{dong2015will,gelli2015image}, and passive-aggressive algorithms~\cite{petrovic2011rt}.
Nonetheless, the application of feature-based methods can pose challenges due to their reliance on domain experts' knowledge, which often restricts the generalizability of learned features to novel contexts. For example, features extracted from tweets may not be directly applicable to scientific papers.

\textbf{Generative-based approaches}: These methods model the underlying diffusion process as event sequences in the continuous temporal domain. The dissemination of information items is commonly characterized using probabilistic generative models, including epidemic models~\cite{matsubara2012rise}, survival analysis~\cite{lee2012modeling}, and various stochastic point processes (e.g., Poisson process~\cite{wang2013quantifying,shen2014modeling}, Hawkes process~\cite{zhao2015seismic,yang2013mixture,zhang2022anytime}).
Lee et al.~\cite{lee2012modeling} borrowed the idea from survival analysis to predict the popularity of online content. 
Shen et al.~\cite{shen2014modeling} proposed a generative probabilistic framework using the reinforced Poisson process to model the arrival process of individual attention. It maximizes the likelihood of the process and obtains popularity via a differential equation. 
SEISMIC~\cite{zhao2015seismic} models the information cascades as a self-exciting point process on Galton-Watson trees during one month of complete Twitter data. 
It imposes no parametric assumptions and avoids expensive feature engineering. 
However, these methods often overlook the implicit diffusion patterns and dynamic characteristics inherent in cascade graphs that play a crucial role in cascade popularity prediction.

\textbf{Deep-learning-based approaches:}
In recent times, the success of deep learning methodologies has triggered the development of numerous information cascade models based on deep neural networks.
DeepHawkes~\cite{cao2017deephawkes} is a both generative and deep-learning-based method that builds the whole network on the theory of the Hawkes process in an end-to-end manner. Hawkesformer~\cite{yu2022transformer} proposes an enhanced neural Hawkes process model with the attention mechanism.
DeepCas~\cite{li2017deepcas} is the first cascade graph representation learning method that distinguishes itself by utilizing DeepWalk for capturing the structural characteristics of information cascades and employing GRU to capture the temporal dynamics of such cascades.
TOPO-LSTM~\cite{wang2017topological} proposes a directed acyclic graph-structured RNN to generate a topology-aware embedding for each user.
Furthermore, there are several similar works: DTCN~\cite{wu2017sequential}, UHAN~\cite{zhang2018user}, and DFTC~\cite{liao2019popularity}, all of which focus on extracting comprehensive diffusion paths from sequential observations of information cascades and employs RNNs for temporal modeling. In addition, there are several other studies~\cite{wang2022casseqgcn,chen2019cascn,zhou2020continual,chen2022multi} that adopt a different approach by employing dynamic graph convolutional networks (GCNs) to learn the structure of each cascade.
Furthermore, VaCas~\cite{zhou2020variational} and CasFlow~\cite{xu2021casflow} propose a hierarchical graph learning method to capture the dynamic evolving cascades graph structure and leverages variational autoencoder or normalizing flows for modeling the uncertainty between the sub-graphs and the whole cascades, with a bi-directional GRU to capture temporal characteristics.
CTCP~\cite{lu2023continuous} integrates multiple cascades into a diffusion graph and considers the correlation between cascades and users' dynamic preferences, incorporating an evolution learning module that continually updates the states of both users and cascades. 
Nevertheless, all the above methods rely on RNNs for temporal pattern learning and thus fail to handle irregular time cascades, which as we show in this paper can significantly improve the prediction performance using our ConCat.

\section{Conclusion}
\label{conclusion}
This paper investigates the problem of modeling continuous-time dynamics of cascades for information popularity prediction. By revisiting the existing cascade popularity prediction methods, we identify the importance of modeling continuous-time dynamics of cascades for popularity prediction, in particular, the dynamics after the last event of a cascade until the observation time. To address this issue, we propose ConCat modeling the \underline{Con}tinuous-time dynamics of \underline{Ca}scades for popularity predic\underline{t}ion. It utilizes neural ODE to model continuous patterns between two events and GRU to incorporate jumps into the neural ODE framework for modeling the discrete events, with neural TPPs further capturing the global trend of cascades, for the ultimate goal of popularity prediction. We conduct extensive experiments to evaluate ConCat on three real-world information cascade datasets. Results show that ConCat achieves superior performance compared to a sizeable collection of state-of-the-art baselines, yielding 2.3\%-33.2\% improvement over the best-performing baselines across the three datasets. In particular, our ablation study provides strong support for our design choices, where the propagation to the unified observation time and neural TPPs for global trends both yield a significant improvement.

In the future, we plan to explore more effective methods to further leverage the characteristics of TPPs beyond the intensity integral for popularity prediction.

\bibliographystyle{IEEEtran}
\bibliography{0-ConCat.bib}


\section{Biography Section}
\vspace{-1cm}
\begin{IEEEbiography}
[{\includegraphics[width=1in,height=1.25in,clip,keepaspectratio]{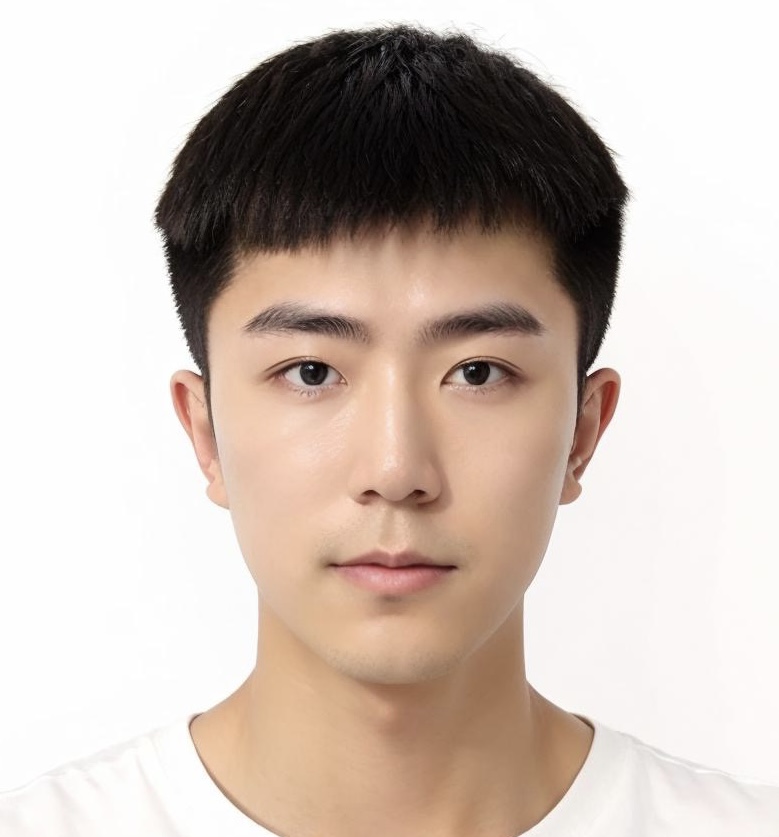}}]{Xin Jing}~ received the BSc and MSc degree both in software engineering from University of Electronic Science and Technology of China. He is currently a PhD student in the Department of Computer and Information Science at University of Macau. His research interests include spatiotemporal data and their applications in various scenarios, primarily on information diffusion in social network and trajectory generation.
\end{IEEEbiography}
\vspace{-0.8cm}
\begin{IEEEbiography}
[{\includegraphics[width=1in,height=1.25in,clip,keepaspectratio]{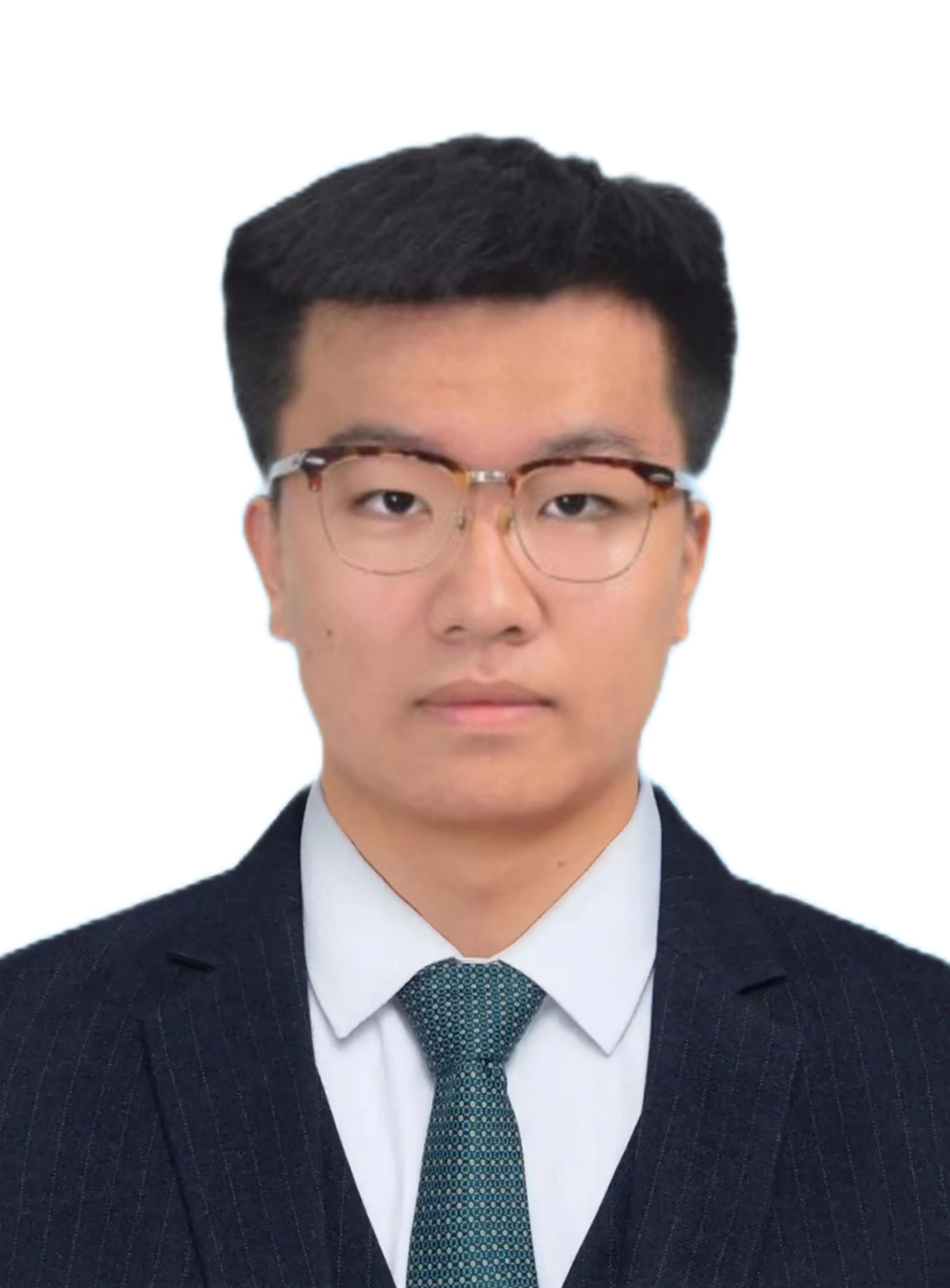}}]{Yichen Jing}~ received the BSc degree in data science and big data technology from Minzu University of China. He is currently a MSc student in the Department of Computer and Information Science at University of Macau. His research mainly focous on temporal data mining.
\end{IEEEbiography}
\begin{IEEEbiography}[{\includegraphics[width=1in,height=1.25in,clip,keepaspectratio]{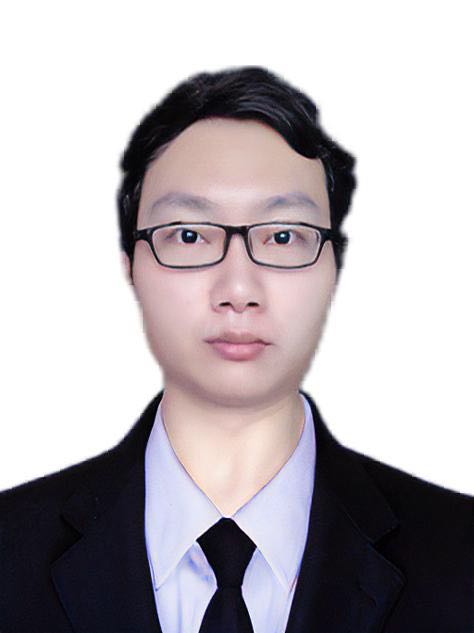}}]{Yuhuan Lu}~ received the BSc and MSc degrees both in transportation engineering from Sun Yat-Sen University. He is currently a PhD student in the Department of Computer and Information Science at University of Macau. His research interests lie in Knowledge Representation, Intelligent Transportation Systems, and Autonomous Driving.
\end{IEEEbiography}
\begin{IEEEbiography}
[{\includegraphics[width=1in,height=1.25in,clip,keepaspectratio]{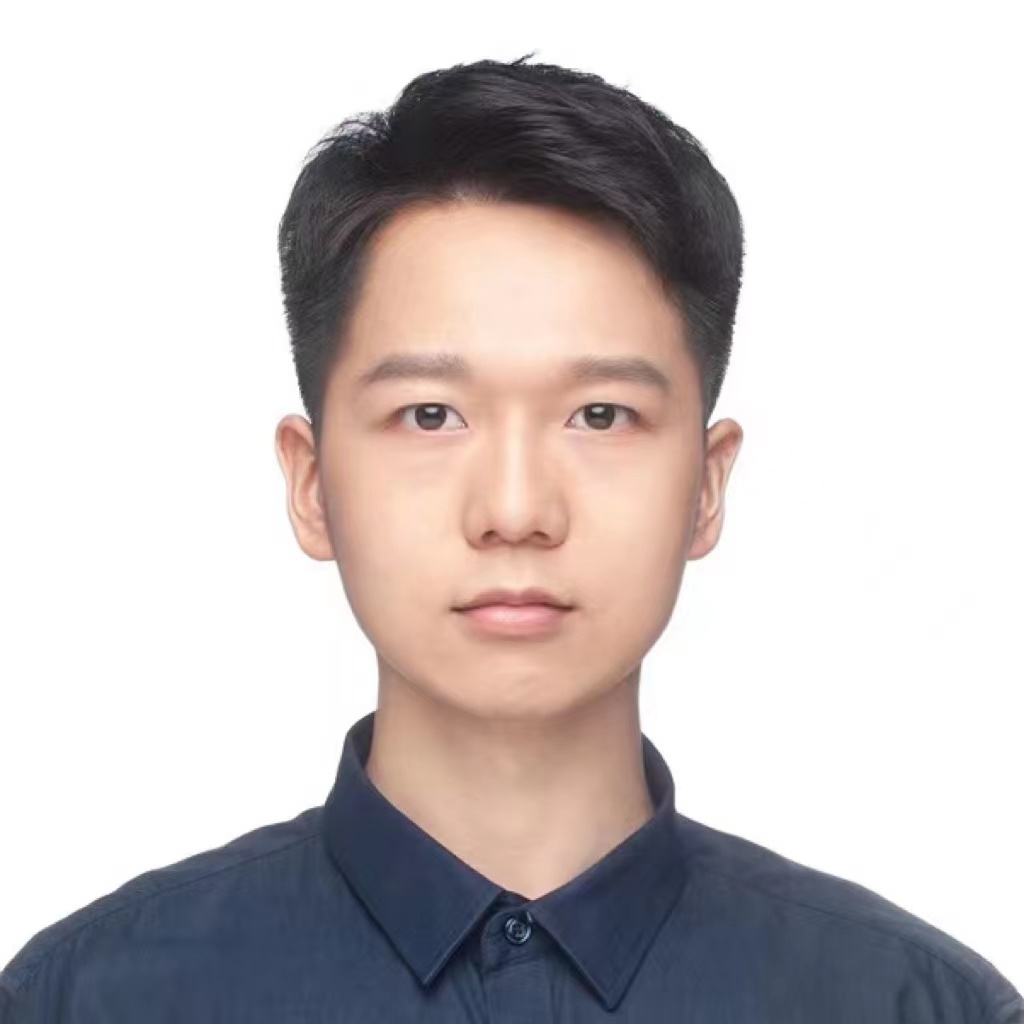}}]{Bangchao Deng}~ received his B.Eng. degree in Computer Science and Technology from Nanjing University of Aeronautics and Astronautics, China. He received his M.S. degree in Computer Science from University of Macau, Macao, China. He is currently a Ph.D. student with the State Key Laboratory of Internet of Things for Smart City and Department of Computer and Information Science, University of Macau, Macao, China. His research interests lie in Spatiotemporal Data Mining and Urban Computing.
\end{IEEEbiography}
\begin{IEEEbiography}
[{\includegraphics[width=1in,height=1.25in,clip,keepaspectratio]{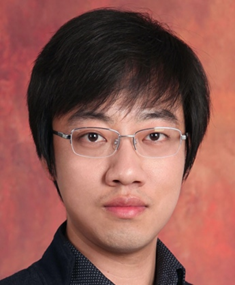}}]{Sikun Yang}~ received the doctoral degree (Dr.-Ing.) from the Technische Universität Darmstadt, Germany. After that, he held postdoctoral positions at Deutsches Zentrum für Neurodegenerative Erkrankungen and The Chinese University of Hong Kong at Shenzhen. Since 2023 he is an assistant professor at the school of computing and information technology at Great Bay University, China. He is interested in probabilistic machine learning, Bayesian inference, stochastic point processes, and their applications.
\end{IEEEbiography}
\begin{IEEEbiography}
[{\includegraphics[width=1in,height=1.25in,clip,keepaspectratio]{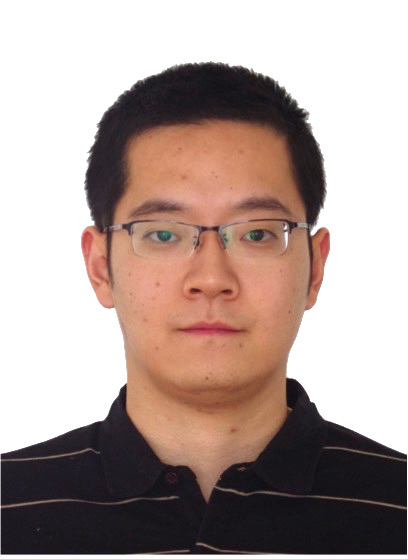}}]{Dingqi Yang}~ is an Associate Professor with the State Key Laboratory of Internet of Things for Smart City and Department of Computer and Information Science, University of Macau. He received his Ph.D. degree in Computer Science from Pierre and Marie Curie University and Institut Mines-TELECOM/TELECOM SudParis in France, where he won both the CNRS SAMOVAR Doctorate Award and the Press Mention in 2015. Before joining the University of Macau, he worked as a senior researcher at the University of Fribourg in Switzerland. His research interests include big data analytics, ubiquitous computing, and smart city.
\end{IEEEbiography}

 




\vfill

\end{document}